\newcommand{\isfulltr}{1}

\ifx\isfulltr\undefined
\documentclass[5p,times,twocolumn]{elsarticle_smi}
\else
\documentclass[5p,times,twocolumn]{elsarticle}
\fi

\usepackage{amsmath}
\usepackage{multirow}
\usepackage{float}

\graphicspath{{./graphics/}}

\begin{document}

%%%%%%%%%%%%%%%%%%%%%%%%%%%%%%%%%%%%%%%%%%%%%

\title{Diffusion-geometric maximally stable component detection in deformable shapes}

\ifx\isfulltr\undefined
\author{Anonymous}
\else
\author{Roee Litman\fnref{tau}}
\author{Alexander M. Bronstein\fnref{tau}}
\author{Michael M. Bronstein\fnref{usi}}
\fntext[tau]{School of Electrical Engineering, Tel Aviv University.}
\fntext[usi]{Institute of Computational Science, University of Lugano.}
\fi

\begin{abstract}
Maximally stable component detection is a very popular method for feature analysis in images, mainly due to its low computation cost and high repeatability.
With the recent advance of feature-based methods in geometric shape analysis, there is significant interest in finding analogous approaches in the 3D world.
In this paper, we formulate a diffusion-geometric framework for stable component detection in non-rigid 3D shapes, which can be used for geometric feature detection and description. A quantitative evaluation of our method on the SHREC'10 feature detection benchmark shows its potential as a source of high-quality features.

%\begin{classification}
% \CCScat{I.3.3}{Computer Graphics}{Picture/Image Generation}{Line and Curve Generation}
%\end{classification}

\end{abstract}

%feature detection; feature descriptor; non-rigid shapes; deformable shapes; maximally stable components; diffusion geometry; intrinsic geometry; level sets; component tree; MSER

\maketitle

\section{Introduction}

Over the past decade, feature-based methods have become a ubiquitous tool in image analysis and a {\em de facto} standard in many computer vision and pattern recognition problems.
More recently, there has been an increased interest in developing similar methods for the analysis of 3D shapes.
Feature descriptors play an important role in many shape analysis applications, such as finding shape correspondence \cite{thorstensen-non}  or assembling fractured models \cite{huang2006reassembling} in computational aracheology.
Bags of features \cite{siv:zis:CVPR:03,ovsjanikov2009shape,TolCasFus09} and similar approaches \cite{mitra2006pfs} were introduced as a way to construct global shape descriptors that can be efficiently used for large-scale shape retrieval.

%(the reader is referred to \cite{bronstein2010shrec_dnd} for a detailed feature detection benchmark)

Many shape feature detectors and descriptors draw inspiration from and follow analogous methods in image analysis.
For example, detection of geometric structures analogous to corners \cite{SB10b} and edges \cite{KolShiTal09} in images has been studied.
The histogram of intrinsic gradients used in \cite{zaharescu-surface} is similar in principle to the scale invariant feature transform (SIFT)  \cite{low:IJCV:04} which has recently become extremely popular in image analysis.
In \cite{gelfand2005rgr}, the integral invariant signatures \cite{manay2004iis} successfully employed in 2D shape analysis were  extended to 3D shapes.

Examples of 3D-specific descriptors include the popular spin image \cite{johnson1999usi},
based on representation of the shape normal field in a local system of coordinates.
Recent studies introduced versatile and computationally efficient descriptors
based on the heat kernel \cite{sun2009concise,Iasonas} describing the local heat propagation properties on a shape. The advantage of these methods is the fact that heat diffusion geometry is intrinsic and thus deformation-invariant, which makes descriptors based on it
applicable in deformable shape analysis.

\subsection{Related work}
A different class of feature detection methods tries to find stable components or regions in the analyzed image or shape.
In the image processing literature, the watershed transform is the precursor of many algorithms for stable component detection \cite{couprie1997topological,vincent2002watersheds}.
In the computer vision and image analysis community, stable component detection is used in the maximally stable extremal regions (MSER) algorithm \cite{matas2004robust}. MSER represents intensity level sets as a component tree and attempts finding level sets with the smallest area variation across intensity; the use of area ratio as the stability criterion makes this approach affine-invariant, which is an important property in image analysis, as it approximates viewpoint transformations.
Alternative stability criteria based on geometric scale-space analysis have been recently proposed in \cite{kimmel-mser}.

In the shape analysis community, shape decomposition into characteristic primitive elements was explored in \cite{mortara2003blowing}.
Methods similar to MSER have been explored in the works on topological persistence \cite{edelsbrunner2002topological}. Persistence-based clustering \cite{chazal2009persistence} was used by Skraba {\em et al.} \cite{skraba2010persistence} to perform shape segmentation.
In \cite{digne2010level}, Digne {\em et al.} extended the notion of vertex-weighted component trees to meshes and proposed to detect MSER regions using
the mean curvature. The approach was tested only in a qualitative way, and not evaluated as a feature detector.

\subsection{Main contribution}
The main contribution of our framework is three-fold.
First, in Section 2 we introduce a generic framework for stable component detection,
which unites vertex- and edge-weighted graph representations
(as opposed to vertex-weighting used in image and shape maximally stable component detectors \cite{matas2004robust,digne2010level}).
Our results (see Section 4) show that the edge-weighted formulation is more versatile and outperforms its vertex-weighted counterpart in terms
of feature repeatability.
Second, in Section 3 we introduce diffusion geometric weighting functions suitable for both vertex- and edge-weighted component trees.
We show that such functions are invariant under a large class of transformations, in particular, non-rigid inelastic deformations, making them especially attractive in non-rigid shape analysis.
We also show several ways of constructing scale-invariant weighting functions.
Third, in Section 4 we show a comprehensive evaluation of different settings of our method on a standard feature detection benchmark
comprising shapes undergoing a variety of transformations (also see Figures~\ref{fig_region_tosca} and \ref{fig_vw_region_sample_trans}).

\section{Diffusion geometry}

Diffusion geometry is an umbrella term referring to geometric analysis of diffusion or random walk processes.
We models a shape as a compact two-dimensional Riemannian manifold $X$.
In it simplest setting, a diffusion process on $X$ is described by the partial differential equation
\begin{eqnarray}
\left(\frac{\partial}{\partial t} + \Delta \right)f(t,x) = 0,
\label{eq:heat}
\end{eqnarray}
called the \emph{heat equation}, where $\Delta$ denotes the positive-semidefinite Laplace-Beltrami operator associated with the Riemannian metric of $X$.
The heat equation describes
the propagation of heat on the surface and its solution $f(t,x)$ is the heat distribution at a point $x$ in time $t$.
The initial condition of the equation is some initial heat distribution $f(0,x)$;
if $X$ has a boundary, appropriate boundary conditions must be added.

%{\bf Heat kernel. }
%
The solution of~(\ref{eq:heat}) corresponding to a point initial condition
 $f(0,x) = \delta(x,y)$, is called the {\em heat kernel} and represents the amount of heat
 transferred from $x$ to $y$ in time $t$ due to the diffusion process.
 The value of the heat kernel $h_t(x,y)$ can also be interpreted as the transition probability
density of a random walk of length $t$ from the point $x$ to the point $y$.

Using spectral decomposition, the heat kernel can be represented as
\begin{eqnarray}
h_t(x,y) &=& \sum_{i\geq 0} e^{-\lambda_i t} \phi_i(x) \phi_i(y).
\label{eq:heatkernel}
\end{eqnarray}
Here, $\phi_i$ and $\lambda_i$ denote, respectively, the eigenfunctions and eigenvalues of the Laplace-Beltrami operator
 satisfying $\Delta \phi_i = \lambda_i \phi_i$ (without loss of generality, we assume $\lambda_i$ to be sorted in increasing order starting with
 $\lambda_0 = 0$).
Since the Laplace-Beltrami operator is an {\em intrinsic} geometric quantity, i.e., it can be expressed solely in terms of the
 metric of $X$, its eigenfunctions and eigenvalues as well as the heat kernel are invariant under isometric transformations (bending) of
 the shape.

The parameter $t$ can be given the meaning of \emph{scale}, and the family $\{ h_t \}_t$ of heat kernels
can be thought of as a scale-space of functions on $X$.
By integrating over all scales, a \emph{scale-invariant} version of~(\ref{eq:heatkernel}) is obtained,
\begin{eqnarray}
c(x,y) &=& \sum_{i\geq 1} \frac{1}{\lambda_i} \phi_i(x) \phi_i(y).
\label{eq:ctkernel}
\end{eqnarray}
This kernel is referred to as the {\em commute-time kernel} and can be interpreted as the transition probability
density of a random walk of any length.

By setting $y=x$, both the heat and the commute time kernels, $h_t(x,x)$ and $c(x,x)$ express the probability density of remaining at a point $x$, respectively after time $t$ and after any time.
The value $h_t(x,x)$, sometimes referred to as the \emph{auto-diffusivity function}, is related to the Gaussian curvature $K(x)$ through
\begin{eqnarray}
h_t(x,x) &\approx & \frac{1}{4\pi t}\left(1 + \frac{1}{6}K(x) t + \mathcal{O}(t^2)\right).
\end{eqnarray}
This relation coincides with the well-known fact
that heat tends to diffuse slower at points with positive
curvature, and faster at points with negative curvature.

For any $t > 0$, the values of $h_t(x,y)$ at every $x$ and $y\in B_\epsilon(x)$ in a small neighborhood around $x$
contain full information about the intrinsic geometry of the shape.
Furthermore, Sun \emph{et al.} \cite{sun2009concise} show that under mild technical conditions, the set $\{ h_t(x,x) \}_{t > 0}$ is also fully informative (note that
the auto-diffusivity function has to be evaluated at all values of $t$ in order to contain full information about the shape metric).
%

%{\bf Diffusion metrics. }
%%
%The heat kernel gives rise to a distance metric referred to as {\em diffusion distance}  and expressed as
%\begin{eqnarray}
%d^2_t(x,x') &=& \sum_{i\geq 1} e^{-\lambda_i t} (\phi_i(x) - \phi_i(x'))^2.
%\end{eqnarray}
%%
%The diffusion distance can be regarded as a ``connectivity'' of $x$ and $y$ by means of random walks of length $t$.
%
%Similarly, the commute-time kernel gives rise to the scale-invariant {\em commute-time metric},
%\begin{eqnarray}
%d^2(x,x') &=& \sum_{i\geq 1} \frac{1}{\lambda_i} (\phi_i(x) - \phi_i(x'))^2,
%\end{eqnarray}
%%
%related to the diffusion distance by the relation
%\begin{eqnarray}
%d^2(x,x') &=& \frac{1}{2}\int_0^\infty d^2_t(x,x') dt.
%\end{eqnarray}

\subsection{Numerical computation}

In the discrete setting, we assume that the shape is sampled at a finite number of points $V = \{ v_1,\hdots,v_N \}$,
upon which a simplicial complex (triangular mesh) with vertices $V$, edges $E \subset V \times V$ and faces $F \subset V \times V \times V$ is constructed.
The computation of the discrete heat kernel $h_t(v_1, v_2)$ and the associated diffusion geometry constructs
is performed using formula~(\ref{eq:heatkernel}), in which a finite number of eigenvalues and eigenfunctions of the discrete Laplace-Beltrami operator are taken.
The latter can be computed directly using the finite elements method (FEM) \cite{reuter},
of by discretization of the Laplace operator on the mesh followed by its eigendecomposition.
Here, we adopt the second approach according to which the discrete Laplace-Beltrami operator is expressed in the following generic form,
\begin{equation}
\label{eq:ldb_discrete}
(\Delta_X f)_i = \frac{1}{a_i} \sum_{j} w_{ij} (f_i - f_j),
\end{equation}
where $f_i = f(v_i)$ is a scalar function defined on $V$, $w_{ij}$ are weights, and $a_i$ are
normalization coefficients.
In matrix notation, (\ref{eq:ldb_discrete}) can be written as
$\Delta_X f = A^{-1} W f$,
where $f$ is an $N\times 1$ vector, $A = \mathrm{diag}(a_i)$ and $W = \mathrm{diag}\left(\sum_{l\neq i} w_{il} \right) - (w_{ij})$.
The discrete eigenfunctions and eigenvalues are found by solving the \emph{generalized eigendecomposition} \cite{levy2006lbe}
$ W \Phi = A \Phi \Lambda$,
where $\Lambda = \mathrm{diag}(\lambda_l)$ is a diagonal matrix of eigenvalues and $\Phi = (\phi_l(v_i))$ is the matrix of the corresponding eigenvectors.

Different choices of $A$ and $W$ have been studied,
depending on which continuous properties of the Laplace-Beltrami operator one
wishes to preserve \cite{floater2005surface,wardetzky2008discrete}. % (see e.g. \cite{zhang2004discrete,floater2005surface,bobenko2007discrete}).
For triangular meshes, a popular choice adopted in this paper is
the \emph{cotangent weight} scheme \cite{pinkall1993computing,meyer2003ddg},
in which
\begin{equation}
w_{ij} = \left\{
\begin{array}{cc}
(\cot \alpha_{ij} + \cot \beta_{ij})/2 & (v_i,v_j) \in E;\\
0 & else,
\end{array}
\right.
\end{equation}
%$j$ in the 1-ring neighborhood of vertex $i$ and zero otherwise,
where $\alpha_{ij}$ and $\beta_{ij}$ are the two
angles opposite to the edge between vertices $v_i$ and $v_j$ in the two triangles sharing the edge,
and $a_i$ are the discrete area elements.

\section{Maximally stable components}

%\subsection{Component trees}
%
Let us now focus on the undirected graph with the vertex set
$V$ and edge set $E$ underlying the discretization of a shape, which with some abuse of notation
we will henceforth denote as $X=(V,E)$.
We say that two vertices $v_1$ and $v_2$ are \emph{adjacent} if $(v_1,v_2) \in E$. An ordered sequence
$\pi = \{v_1,\dots,v_k\}$ of vertices is called a \emph{path} if for any $i=1,\dots,k-1$, $v_i$ is adjacent to $v_{i+1}$. In this case, we say
that $v_1$ and $v_k$ are linked in $X$. The graph is said to be \emph{connected} if every pair of vertices in it is linked.
A graph $Y=(V'\subseteq V,E' \subseteq E)$ is called a \emph{subgraph} of $X$ and denoted by $Y \subseteq X$. We say that $Y$ is a (connected) \emph{component} of
$X$ if $Y$ is a connected subgraph of $X$ that is maximal for this property (i.e., for any connected subgraph $Z$,  $Y \subseteq Z \subseteq X$ implies
$Y = Z$).
Given $E' \subseteq E$, the graph induced by $E'$ is the graph $Y = (V',E')$ whose vertex set is made of all vertices belonging to an edge in $E'$, i.e.,
$V' = \{ v \in V : \exists v' \in V, (v,v') \in E' \}$.

A scalar function $f : V \rightarrow \mathbb{R}$ is called a \emph{vertex weight}, and a graph equipped with it is called \emph{vertex-weighted}. Similarly, a graph equipped with a  function
$d : E \rightarrow \mathbb{R}$ defined on the edge set is called \emph{edge-weighted}.
In what follows, we will assume both types of weights to be non-negative.
Grayscale images are often represented as vertex-weighted graphs with some regular (e.g., four-neighbor) connectivity and weights
corresponding to the intensity of the pixels. Edge weights can be obtained, for example, by considering a local distance function measuring the dissimilarity of pairs of adjacent pixels. While vertex weighting is limited to scalar (grayscale) images, edge weighting is more general.
%

%{\bf Component trees. }
\subsection{Component trees}
Let $(X,f)$ be a vertex-weighted graph. For $\ell \ge 0$, the $\ell$-\emph{cross-section} of $X$ is defined as the graph induced by $E_\ell = \{ (v_1,v_2) \in E : f(v_1), f(v_2) \le \ell \}$. Similarly, a cross-section of an edge-weighted graph $(X,d)$ is induced by the edge subset
$E_\ell = \{ e \in E : d(e) \le \ell \}$.
A connected component of the cross-section is called an $\ell$-level set of the weighted graph.
%Since the definition is valid for both vertex- and edge-weighted graphs, cross-sections will be denoted simply by $X_\ell$ whenever possible.

For any component $C$ of $X$, we define the \emph{altitude} $\ell(C)$ as the minimal $\ell$ for which $C$ is a
component of the $\ell$-cross-section of $X$. Altitudes establish a partial order relation on the connected components of $X$
as any component $C$ is contained in a component with higher altitude.
The set of all such pairs $(\ell(C), C)$ therefore forms a tree called the \emph{component tree}. Note that the above definitions are valid for
both vertex- and edge-weighted graphs.

%\subsection{Maximally stable components}

%{\bf Maximally stable components. }
\subsection{Maximally stable components}
Since in our discussion undirected graphs are used as a discretization of smooth manifolds, we can associate with every component $C$ (or every subset of the vertex set in general) a measure of \emph{area}, $A(C)$. In the simplest setting, the area of $C$ can be thought of as its cardinality. In a better discretization, each vertex $v$ in the graph is associated with a discrete area element $da(v)$, and the area of a component is defined as
\begin{eqnarray}
A(C) &=& \sum_{v \in C} da(v).
\end{eqnarray}

Let now $\{ (\ell,C_\ell) \}$ be a sequence of nested components forming a branch in the component tree.
We define the \emph{instability} of $C_\ell$ as
\begin{eqnarray}
s(\ell) &=& \frac{dA(C_\ell)}{d\ell}.
\label{eq:stability}
\end{eqnarray}
In other words, the more the area of a component changes with the change of $\ell$, the less stable it is.
A component $C_{\ell^*}$ is called \emph{maximally stable} if the instability function has
a local minimum at $\ell^*$.
Maximally stable components are widely known
in the computer vision literature under the name of \emph{maximally stable extremal regions} or \emph{MSERs} for short \cite{matas2004robust}, with $s(\ell^*)$ usually referred to as the region \emph{score}.

It is important to note that in their original definition, MSERs were defined on a component tree of a vertex-weighted graph, while
our definition is more general and allows for edge-weighted graphs as well.
The importance of such an extension will become evident in the sequel.
Also, the original MSER algorithm \cite{matas2004robust} assumes the vertex weights to be quantized, while our formulation is suitable for scalar fields whose dynamic range is unknown \emph{a priori}.

\subsection{Computational aspects}

We use the quasi-linear time algorithm detailed in \cite{najman2006building} for the construction of vertex-weighted component trees, and its
straightforward adaptation to the edge-weighted case. The algorithm is based on the observation that the vertex set $V$ can be partitioned
into disjoint sets which are merged together as one goes up in the tree. Maintaining and updating such a partition can be performed very efficiently
using the \emph{union-find} algorithm and related data structures. The resulting tree construction complexity is $\mathcal{O}(N \log \log N)$.

The derivative (\ref{eq:stability}) of the component area with respect to $\ell$ constituting the stability function is computed using finite differences in each branch of the tree.
For example, in a branch $C_{\ell_1} \subseteq C_{\ell_2} \subseteq \cdots \subseteq C_{\ell_K}$,
\begin{eqnarray}
s(\ell_k) &\approx& \frac{A(C_{\ell_{k+1}}) - A(C_{\ell_{k-1}})}{\ell_{k+1}-\ell_{k-1}}.
\end{eqnarray}
The function is evaluated and its local minima are detected in a single pass over the branches of the component tree starting from the leaf nodes.
We further filter out maximally stable regions with too high values of $s$. In cases where two nested regions overlapping by more that a predefined threshold are detected as maximally stable, only the bigger one is kept.

\section{Weighting functions}

Unlike images where methods based on the analysis of the component tree have been shown to be extremely successful e.g. for segmentation
or affine-invariant feature detection (namely, the MSER feature detector), similar techniques have been only scarcely explored for 3D shapes (with the notable exceptions
 of \cite{digne2010level} and \cite{skraba2010persistence}).
One of possible reasons is the fact that while images readily offer pixel intensities as the trivial vertex weight field,
3D shapes are not generally equipped with any such field. While the use of the mean curvature was proposed in \cite{digne2010level}, it lacks most of invariance
properties required in deformable shape analysis.
%
%\subsection{Diffusion geometry}
%
Here, we follow \cite{skraba2010persistence} in adopting the diffusion geometry framework and show that it allows to construct both vertex and edge weights suitable for the definition of maximally stable components with many useful properties.

%\subsection{Vertex weights}

Given a vertex $v$, the values of the discrete auto-diffusivity function can be directly used as the vertex weights,
\begin{eqnarray}
f(v) &=& h_t(v,v).
 \label{eq:vw-htxx}
\end{eqnarray}
Maximally stable components defined this way
are intrinsic and, thus, invariant to non-rigid bending.
Such strong invariance properties are particularly useful in the analysis of deformable shapes.
However, unlike images where the intensity field contains all information about the image, the above weighting function does not describe the intrinsic geometry of the shape entirely. It furthermore depends on the selection of the scale parameter $t$.

%On the other hand, for every $t > 0$, the function $h_t(x,y)$ for every $x$ and $y$ in some small neighborhood of $x$ contains full information about the intrinsic geometry. Furthermore, Sun \emph{et al.} \cite{sun2009concise}  show that under mild technical conditions, $\{ h_t(x,x) \}_{t > 0}$ is also fully informative.
%%

%\subsection{Edge weights}

Edge weights constitute a more flexible alternative allowing to incorporate fuller geometric information.
The simplest edge weighting scheme can be obtained from a vector-valued field defined on the vertices of the graph.
For example, associating $h_t(v,v)$ for $t\in [t_1,t_2]$ with each vertex $v$, one can define an edge weighting function
\begin{eqnarray}
\lefteqn{d(v_1,v_2)  \ = \ \| h_t(v_1,v_1) - h_t(v_2,v_2) \|_t}  \nonumber\\
&& =\ \left( \int_{t_1}^{t_2} (h_t(v_1,v_1) - h_t(v_2,v_2) )^2 dt \right)^{1/2}
 \label{eq:ew-norm}
\end{eqnarray}
(here, we write $\| \cdot \|_t$ to make explicit that the norm is taken with respect to the variable $t$).
The function has a closed-form expression that can be obtained by substituting the spectral decomposition (\ref{eq:heatkernel}) of the heat kernel.
The advantage of this approach stems from its ability to incorporate multiple scales. Theoretically, the set of $h_t(v,v)$ at all scales contains full information about
the intrinsic geometry of the shape.

%As edge weight, we can use either some notion of distance between the edge vertices $(v_1,v_2) \in E$.
%%
%One possibility is to use the diffusion or the commute-time distance.
%
In a more general setting, edge weights do not necessarily need to stem from any finite- or infinite-dimensional vector field defined on the vertices.
For example, since the discrete heat kernel $h_t(v_1,v_2)$ represents ``proximity'' between $v_1$ and $v_2$, a function
inversely proportional to the value of the heat kernel, e.g.
\begin{eqnarray}
d(v_1,v_2) &=& \frac{1}{h_t(v_1,v_2)}
 \label{eq:ew-htxy}
\end{eqnarray}
can be used as an edge weight. For sufficiently small values of $t$, this function also contains full information about the shape's intrinsic geometry.

Another way of creating edge weights inversely proportional to $h_t$ is by integrating the squared difference between the kernels centered at $v_1$ and $v_2$ over the entire shape,
\begin{eqnarray}
\lefteqn{d(v_1,v_2)  \ = \ \| h_t(v_1,\cdot) - h_t(v_2,\cdot) \|_X}  \nonumber\\
&& =\ \left( \sum_{v \in V} (h_t(v_1,v) - h_t(v_2,v) )^2 da(v) \right)^{1/2}.
 \label{eq:ew-diffusion}
\end{eqnarray}
This construction has been previously introduced in \cite{coifman2006diffusion} under the name of \emph{diffusion distance}, which constitutes an
intrinsic metric on $X$ and is fully informative for small $t$'s.

%Finally, if scale invariance is further required, the Fourier transform magnitude of the heat kernel $h_t(v_1,v_2)$ can be used, as described in the next section.

\subsection{Scale invariance}

The vertex weighting function (\ref{eq:vw-htxx}) and the edge weighting functions (\ref{eq:ew-norm}), (\ref{eq:ew-htxy}) and (\ref{eq:ew-diffusion}) based on the heat kernel
are not scale invariant since a global scaling of the shape by a factor $\gamma > 0$ influences
the heat kernel as $\gamma^2 h_{\gamma^2 t}(v_1,v_2)$, scaling by $\gamma^2$ both the time parameter and the kernel itself.
A possible remedy is to replace the heat kernel by the scale invariant commute time kernel. However, due to the slow decay of the expansion coefficients $\lambda_i^{-1}$ in (\ref{eq:ctkernel}) compared to $e^{-\lambda_i t}$ in (\ref{eq:heatkernel}), the numerical computation
of the commute time kernel is more difficult as it requires many more eigenfunctions of the Laplacian to achieve the same accuracy.

As an alternative, it is possible to use a sequence of transformations of $h_t(v_1,v_2)$ that renders it scale invariant \cite{Iasonas}.
First, the heat kernel is sampled logarithmically in time.
Next, the logarithm and a derivative with respect to time of the heat kernel values are taken to undo the multiplicative constant.
Finally, taking the magnitude of the Fourier transform allows to undo the scaling of the time variable.
This yields the \emph{modified heat kernel} of the form
\begin{eqnarray}
\hat{h}_\omega (v_1,v_2) &=& \left| \mathcal{F}\left\{ \frac{\partial \log h_{t}(v_1,v_2) }{\partial \log t} \right\}(\omega) \right|,
\label{eq:hFourier}
\end{eqnarray}
where $\omega$ denotes the frequency variable of the Fourier transform. The transform is computed numerically using the FFT as detailed in \cite{Iasonas}.
Substituting $\hat{h}_\omega$ into (\ref{eq:ew-norm})--(\ref{eq:ew-htxy}) yields scale invariant edge weighting functions.\footnote{Since the component inclusion relations giving rise to component tree are invariant to any monotonous transformation of the weighting functions, it is sufficient to undo just undoing the scaling of the time parameter $t$ without undoing the scaling of the kernel itself. However, such a transformation affects the scores of the detected regions. We found that
the logarithmic transformation and derivative improve repeatability. Furthermore, by completely undoing the effect of scaling, the modified heat kernel can be used
both in the weighting function and in descriptors of the maximally stable components as detailed in the following section. }
%
%{\bf [TODO: consider adding explanation of the two function (or drop one / both of them)]}
%
By selecting a single frequency $\omega$, one can construct a scale invariant vertex weight $f(v) = \hat{h}_\omega(v,v)$ similar to (\ref{eq:vw-htxx}).
Another way of constructing a scale invariant vertex weight is by integrating $\hat{h}_\omega$ over a rage of frequencies, e.g.,
\begin{eqnarray}
f(v)  & = & \| \hat{h}_\omega(v,v) \|_\omega =  \left( \int_{\omega_1}^{\omega_2} \hat{h}^2_\omega(v,v) d\omega \right)^{1/2}.
 \label{eq:vw-norm}
\end{eqnarray}

\begin{figure*}[tb]
\begin{small}
    \begin{center}
    \includegraphics[width=.9\linewidth,bb=0 0 2243 564]{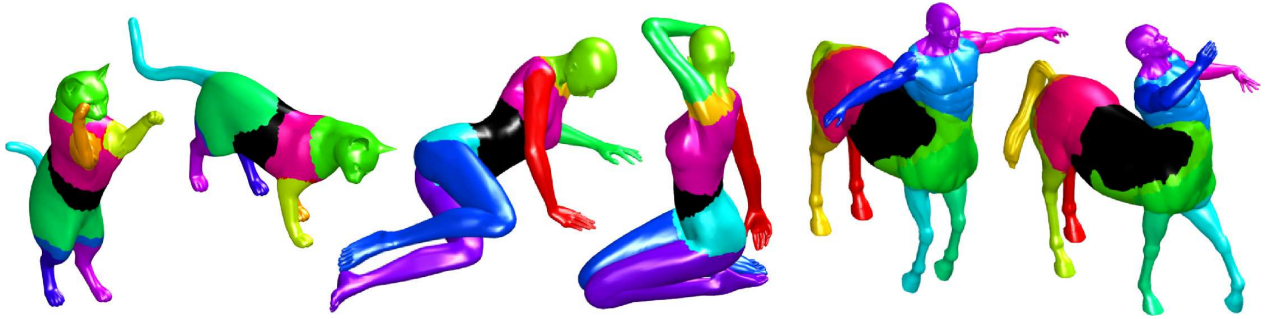}
    \end{center}
\end{small}
\caption{Maximally stable regions detected on different shapes from the TOSCA dataset. Note the invariance of the regions to strong non-rigid deformations.
Also observe the similarity of the regions detected on the female shape and the upper half of the centaur (compare to the male shape from Figure~\ref{fig_vw_region_sample_trans}).
Regions were detected using $h_t(v,v)$ as vertex weight function, with  $t=2048.$ }
\label{fig_region_tosca}
\end{figure*}

% ------------------------------------------------------
\begin{figure*}[tb]
\begin{small}
    \begin{center}
            \includegraphics[width=.9\linewidth,bb=0 0 2595 1512]{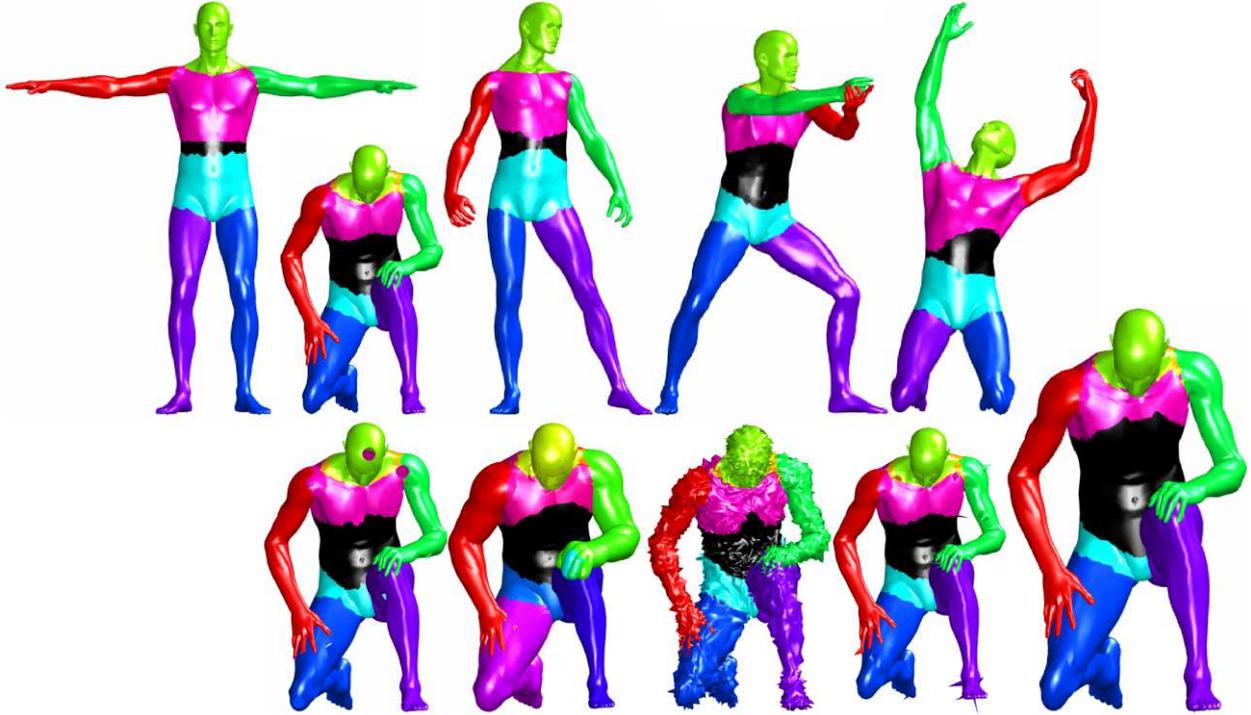}
    \end{center}
\end{small}
\caption{Maximally stable regions detected on shapes from the SHREC'10 dataset using the vertex weight $h_t(v,v)$ with  $t=2048$.
First row: different approximate isometries of the human shape. Second row: different transformations (left-to-right): holes, localscale, noise, shotnoise and scale.}
\label{fig_vw_region_sample_trans}
\end{figure*}

\ifx\isfulltr\undefined
\else
\begin{figure*}[tb]
\begin{small}
    \begin{center}
            \includegraphics[width=\linewidth,bb=0 0 2838 861]{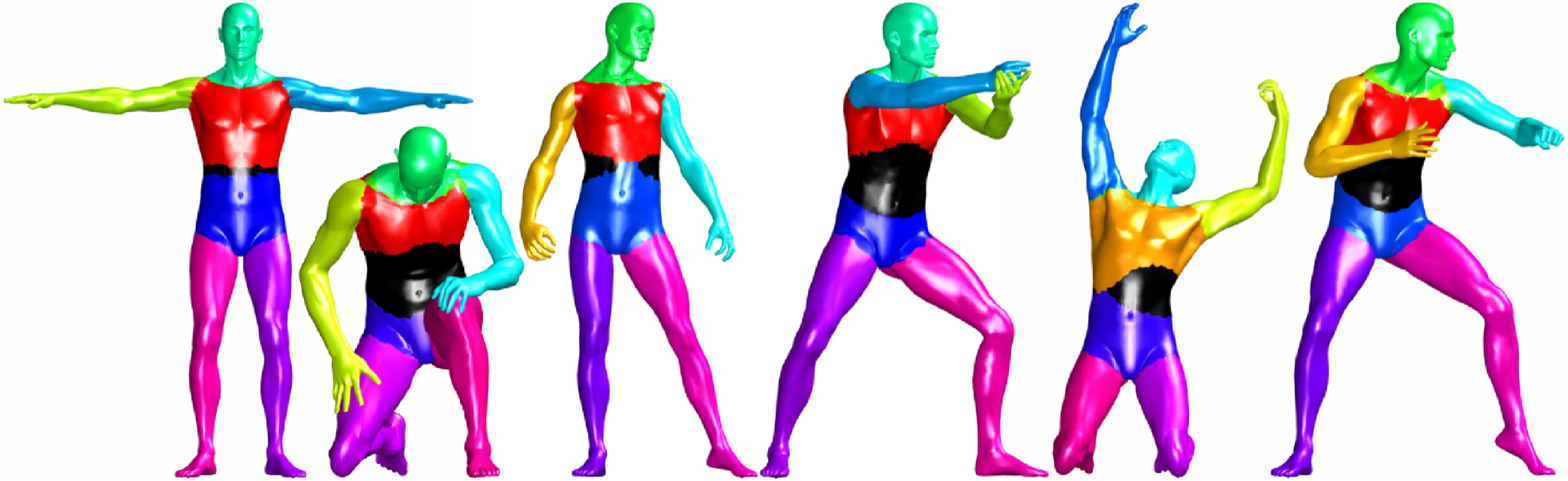}
    \end{center}
\end{small}
\caption{Maximally stable regions detected on shapes from the SHREC'10 dataset using the edge weight $1/h_t(v_1,v_2)$ with  $t=2048$.
Region coloring is arbitrary. }
\label{fig_ew_region_sample_iso}
\end{figure*}
\fi

\section{Descriptors}

\subsection{Point descriptors}

Once the regions are detected, their content can be described using any standard point-wise descriptor of the form $\alpha : V \rightarrow \mathbb{R}^q$.
%
%we sought to attach a descriptor to each of them. Descriptors must represent the region repetitively, and still be discriminative.
%
%{\bf Heat kernel signature. }
In particular, here we consider point-wise {\em heat kernel descriptors} proposed in \cite{sun2009concise}.
The heat kernel descriptor (or {\em heat kernel signature}, HKS) is computed by taking the values of the discrete auto-diffusivity function at vertex $v$ at multiple times, $\alpha(v) = (h_{t_1}(v,v),\hdots, h_{t_q}(v,v))$, where $t_1,\hdots, t_q$ are some fixed time values.
Such a descriptor is a vector of dimensionality $q$ at each point.
Since the heat kernel is an intrinsic quantity, the HKS is invariant to isometric transformations of the shape.
 %
% In the experiments performed in this study, the value of  $q = 6$ was used, and $t_1,\hdots,t_6$ were chosen as $1024, 1351, 1783, 2353, 3104$ and $4096$ (these are settings identical to \cite{ovsjanikov2009shape}).

%{\bf Scale-invariant heat kernel signature. }
%The drawback of the HKS is its dependence of the form $s^2 h_{s^2 t}(v,v)$ on the scaling of the shape by the factor of $s$.
%A solution to this problem was proposed in \cite{Iasonas} using a sequence of HKS transformations that render it scale-invariant.

A scale-invariant version of the HKS descriptor (SI-HKS) can be obtained as proposed \cite{Iasonas} by replacing $h_t$ with
$\hat{h}_\omega$ from (\ref{eq:hFourier}), yielding $\alpha(v) = (\hat{h}_{\omega_1}(v,v),\hdots, \hat{h}_{\omega_q}(v,v))$, where $\omega_1,\dots,\omega_q$
are some fixed frequency values.
%
%%%
%%First, the heat kernel is sampled logarithmically. Next, the logarithm and a derivative with respect to time of the heat kernel values are taken to undo the multiplicative constant. Finally, taking the magnitude of the Fourier transform allows to undo the scaling of the time variable.
%%
%The final descriptor has the form
%\begin{eqnarray}
%\alpha(v) &=& \left\{ \left| \mathcal{F}\left\{ \frac{\partial \log h_{t}(v,v) }{\partial \log t} \right\}(\omega_k) \right| \right\}_{k=1}^r,
%\end{eqnarray}
% %
% where $\omega_1,\hdots,\omega_r$ are some frequencies.
% %
In the following experiments, the heat kernel was sampled at time values $t = 2^1,2^{1+1/16},\hdots,2^{25}$.
The first six discrete frequencies of the Fourier transform were taken, repeating the settings of \cite{Iasonas}.

\subsection{Region descriptors}

Given a descriptor $\alpha(v)$ at each vertex $v \in V$, the simplest way to
define a {\em region descriptor} of a component $C \subset V$ is by
computing the average of $\alpha$ in $C$,
\begin{eqnarray}
\beta(C) = \sum_{v \in C}\alpha(v) da (v).
\label{eq:region-desc-avg}
\end{eqnarray}
The resulting region descriptor $\beta(C)$ is a vector of the same dimensionality $q$ as the point descriptor $\alpha$.

An alternative construction considered here follows Ovsjanikov {\em et al.} \cite{ovsjanikov2009shape} where a
global shape descriptors were obtained from point-wise descriptors
using the {\em bag of features} paradigm \cite{siv:zis:CVPR:03}.
In this approach, a fixed ``geometric vocabulary'' $\alpha_1, \hdots, \alpha_p$ is computed by means of an off-line clustering of the descriptor space.
Next, each point descriptor at $v$ is represented in the vocabulary using vector quantization, yielding a point-wise $p$-dimensional distribution of the form
\begin{eqnarray}
\theta(v) \propto e^{-\| \alpha(v) - \alpha_l \|^2 / 2\sigma^2}.
\end{eqnarray}
The distribution is normalized in such a way that the elements of $\theta(v)$ sum to one.
In the case of $\sigma=0$, hard vector quantization is used, and $\theta_l(v) = 1$ for $\alpha_l$ being the closest element of the
geometric vocabulary to $\alpha(v)$ in the descriptor space, and zero elsewhere.
%
%The bag of features is given as the distribution of geometric words over the entire shape.
%
Given a component $C$, we can define a {\em local bag of features} by computing the distribution of geometric words over the region,
\begin{eqnarray}
\beta(C) = \sum_{v\in C} \theta(v) da(v).
\label{eq:region-desc-bof}
\end{eqnarray}
Such a bag of features is used as a region descriptor of dimensionality $p$.

\section{Results}

\subsection{Dataset}

The proposed approach was tested on the
data of the SHREC'10 feature detection and description benchmark \cite{bronstein2010shrec_dnd}.
The SHREC dataset consisted of three shape classes, with simulated transformations
applied to them.
Shapes are represented as triangular meshes with approximately
10,000 to 50,000 vertices. In our experiments, all meshes were downsampled to at most 10,000 vertices.
Each shape class contained nine categories of transformations: \emph{isometry}
(non-rigid almost inelastic deformations), \emph{topology} (welding of shape vertices resulting in different triangulation), \emph{micro holes} and \emph{big holes} simulating missing data and occlusions, \emph{global} and \emph{local scaling}, \emph{additive Gaussian
noise}, \emph{shot noise}, and \emph{downsampling} (less than 20\% of the original points).
In transformation appeared in five different strengths.
Vertex-wise correspondence between the transformed and the null shapes was given and used as the ground truth in the evaluation of region detection repeatability.
Since all shapes exhibit intrinsic bilateral symmetry, best results over the groundtruth correspondence and its symmetric counterpart were used.
%As some of the transformed shapes had missing data compared to the null shape,
%comparison has to be defined in a one-way manner.
%When ever regions of two shapes are said to be compared, only regions in the transformed shape that have no null-region counterpart will decrease the score, while such unmatched null-region will not.

We also used several deformable shapes from the TOSCA dataset \cite{bronstein2008numerical} for a qualitative evaluation.

\subsection{Detector repeatability}

\begin{figure*}[tb]
\begin{small}
    \begin{center}
    \includegraphics[width=\linewidth,bb=0 0 1519 784]{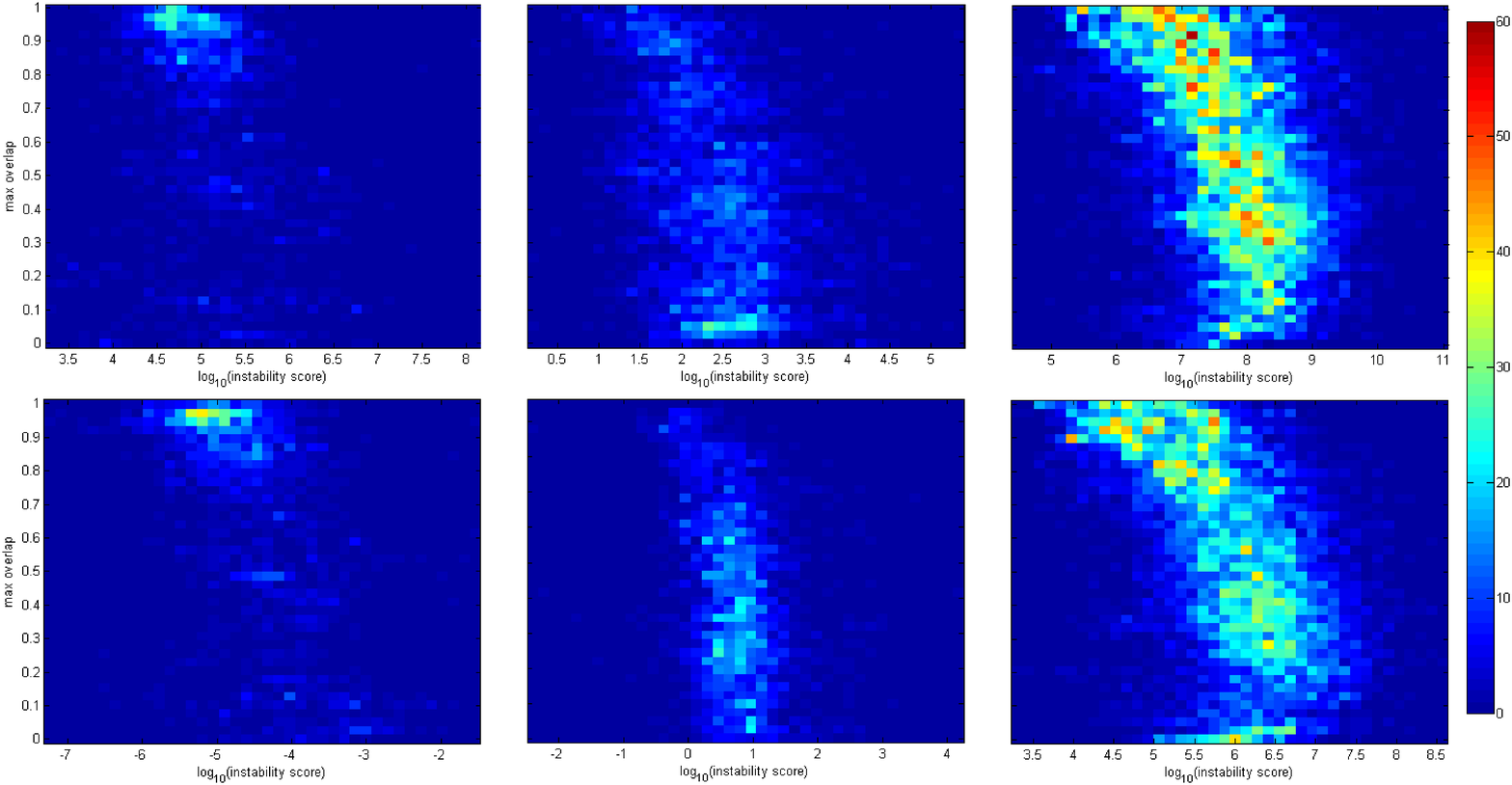}
    \end{center}
\end{small}
\caption{Distributions of maximally stable components as a function of the overlap to the groundtruth regions and instability score.
Left-to-right top-to-bottom are shown the following weighting function: vertex weight $h_t(v,v)$ at $t=2048$, vertex weight $c(v,v)$, edge weight
$\| h_t(v_1,\cdot) - h_t(v_2,\cdot)\|_X$ (diffusion-distance) at $t=2048$, edge weight $1/h_t(v_1,v_2)$ at $t=2048$, vertex weight $\hat{h}_\omega(v,v)$ at $\omega = 0$  and edge weight $| h_t(v_1,v_1) - h_t(v_2,v_2)|$ at $t=2048$.
%
%Left column -  upper: $h_t(v,v)$  lower: $1/h_t(v1,v2)$ %"The good" - weights that needn't a filter, as most detected regions are with high overlap.
%Center column -  upper: $.$  lower: $.$ %"The bad" - weights that filter can't improve them, as most detected regions are with the same score overlap.
%Right column -  upper: $.$  lower: $.$ %"The ugly" - weights that can be improved by a filter, up to a certain score regions are with big overlap.
Good detectors are characterized by a large number of high-overlap stable regions (many regions in the upper left corner of the plot) that can be easily separated from the low-overlap regions that should be concentrated in the lower right corner. }
\label{fig_histograms}
\end{figure*}

The evaluation of the proposed feature detector and descriptor followed the spirit of the influential work by Mikolajczyk \emph{et al.} \cite{mikolajczyk2005comparison}.
In the first experiment, the repeatability of the detector was evaluated.
Let $X$ and $Y$ be the null and the transformed version of the same shape, respectively.
Let $X_1,\hdots, X_m$ and $Y_1,\hdots, Y_n$ denote the regions detected in $X$ and $Y$, and let $X'_j$ be the image of the region $Y_j$ in $X$ under the ground-truth correspondence.\footnote{As some of the transformed shapes had missing data compared to the null shape,
comparison was defined single-sidedly.
Only regions in the transformed shape that had no corresponding regions in the null counterpart decreased the overlap score, while unmatched regions of the null shape did not.}
Given two regions $X_i$ and $Y_j$, their {\em overlap} is defined as the following area ratio
\begin{eqnarray}
\hspace{-5mm} O(X_i,X'_j) = \frac{A(X_i \cap X'_j)}{A(X_i \cup X'_j)} = \frac{A(X_i \cap X'_j)}{A(X_i) + A(X'_j) - A(X_i \cap X'_j)}.
\end{eqnarray}
The {\em repeatability at overlap $o$} is defined as the percentage of regions in $Y$ that have corresponding counterparts in $X$ with overlap greater than $o$ \cite{mikolajczyk2005comparison}.
An ideal detector has the repeatability of $1$.

Four vertex weight functions were compared: discrete heat kernel (\ref{eq:vw-htxx}) with $t=2048$, commute time kernel (\ref{eq:ctkernel}), modified heat kernel with $\omega=0$, and the norm of the modified heat kernel (\ref{eq:vw-norm}).
These four scalar fields were also used to construct edge weights according to $d(v_1,v_2) = |f(v_1)-f(v_2)|$.
Furthermore, since these kernels are functions of a pair of vertices, they were used to define edge weights according to (\ref{eq:ew-htxy}).
In addition, we also tested edge weights constructed according to (\ref{eq:ew-norm}) and the diffusion distance (\ref{eq:ew-diffusion}). Unless mentioned otherwise, $t=2048$ was used for the heat kernel and $\omega= 0$ for the modified heat kernel, as these settings turned out to give best performance on the SHREC'10 dataset.

\ifx\isfulltr\undefined
\else
\begin{figure}[tb]
    \includegraphics[width=\columnwidth,bb=0 0 872 547]{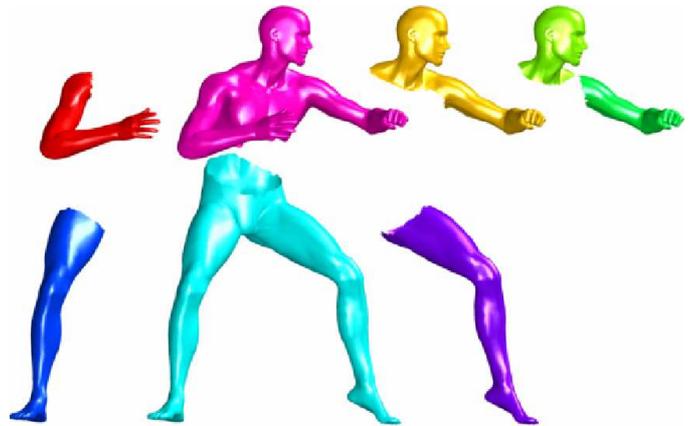}
    \caption{The set of maximally stable regions extracted from one of the shapes in Figure \ref{fig_vw_region_sample_trans}.}
    \label{fig_region_parts}
\end{figure}
\fi

We first evaluated different region detectors qualitatively using shapes from the SHREC'10 and the TOSCA datasets.
Figure~\ref{fig_region_tosca} shows the regions  detected using the vertex weight $h_t(v,v)$ with $t=2048$ on a few sample shapes from the TOSCA dataset.
\ifx\isfulltr\undefined
Figure \ref{fig_vw_region_sample_trans} depicts
\else
Figures \ref{fig_vw_region_sample_trans} and \ref{fig_region_parts} depict
\fi
the maximally stable components detected with the same settings on several shapes from the SHREC dataset.
\ifx\isfulltr\undefined
Similar regions were obtained
\else
Figure~\ref{fig_ew_region_sample_iso} shows the regions obtained
\fi
using the edge weighting function $1/h_t(v_1,v_2)$.
In all cases, the detected regions appear robust and repeatable under the transformations. Surprisingly, many of these regions
have a clear semantic interpretation. Moreover, similarly looking regions are detected on the male and female shapes, and the upper half of the centaur. This makes the proposed feature detector a good candidate for partial shape matching and retrieval.

%{\bf [TODO:  ADD FIGURE WITH TRANSFORMATIONS]}

In order to select optimal cutoff threshold of the instability function (i.e., the maximum region instability value that is still accepted by the detector),
we estimated the empirical distributions of the detecting regions as a function of the instability score and their overlap with the corresponding groundtruth regions.
These histograms are depicted in Figure~\ref{fig_histograms}. An good detector should produce many regions with overlap close to $100\%$ that have low instability, and produce as few as possible low-overlap regions that have very high instability that can be separated from the high-overlap regions by means of a threshold.
In each of the tested detectors, the instability score threshold was selected to maximize the detection of high-overlap regions.

\begin{figure*}[t]
    \includegraphics[width=\linewidth]{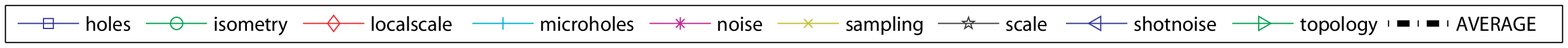} \\
    \begin{tabular}{cc}
        \includegraphics[width=\columnwidth]{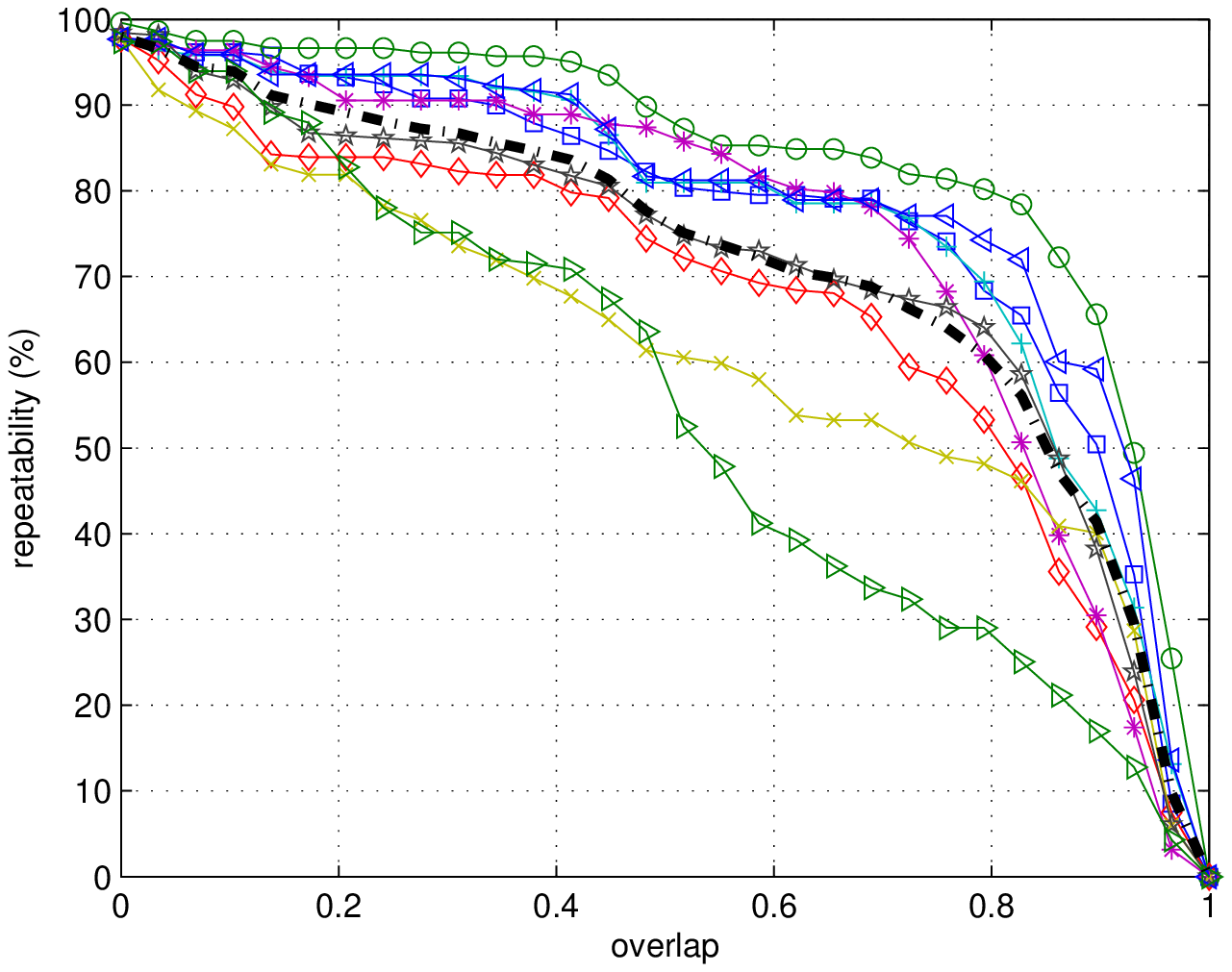} &
        \includegraphics[width=\columnwidth]{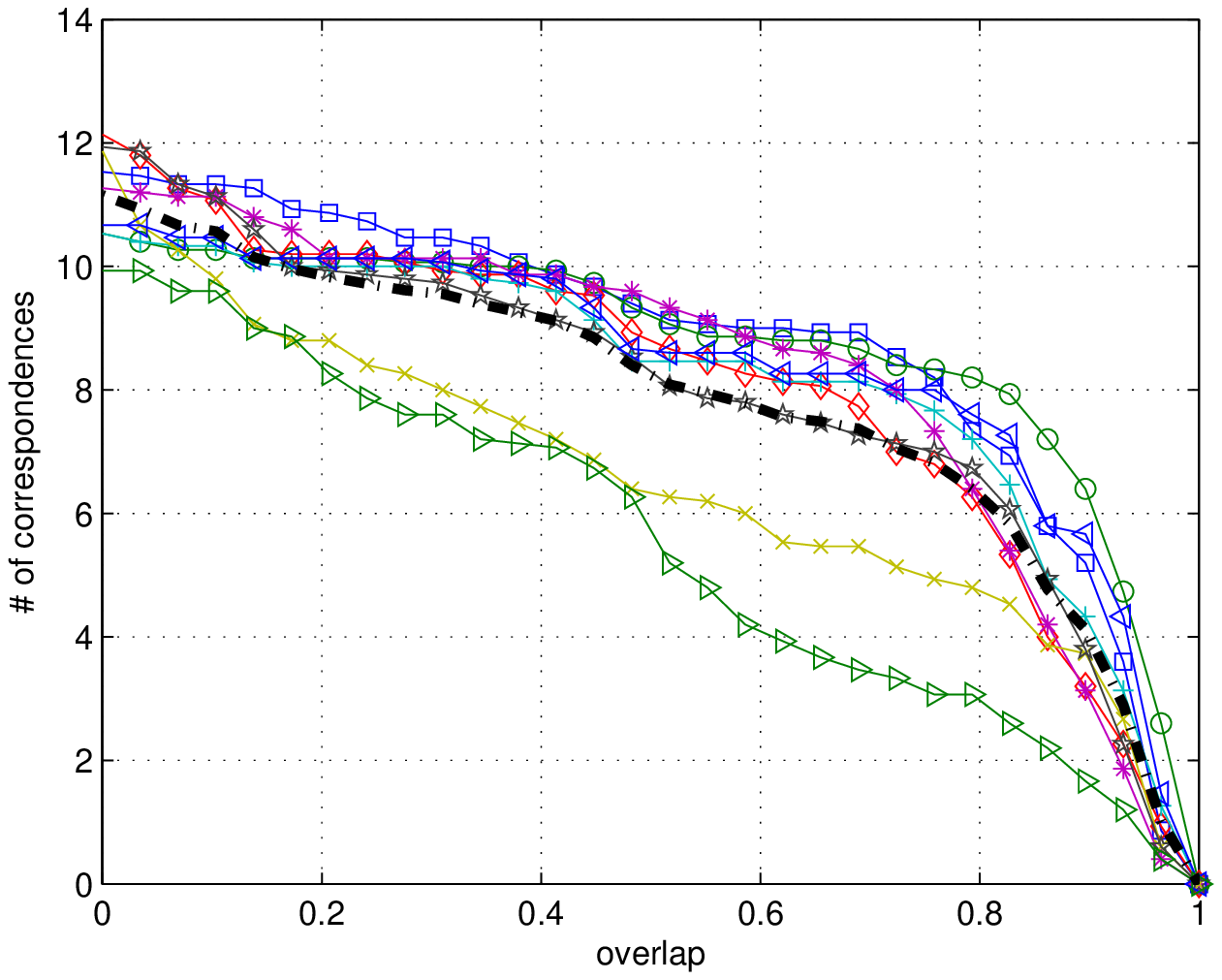} \\

        \includegraphics[width=\columnwidth]{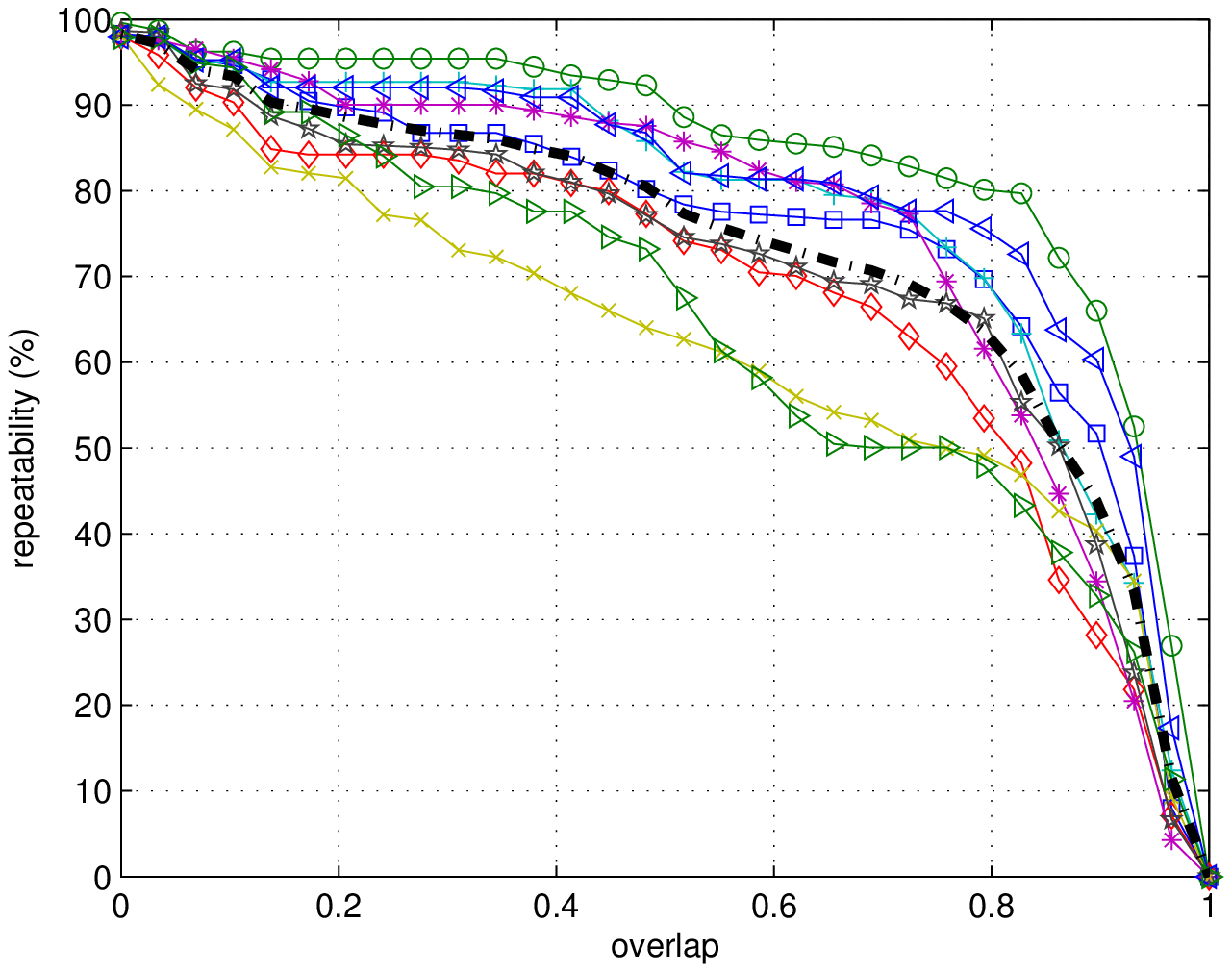} &
        \includegraphics[width=\columnwidth]{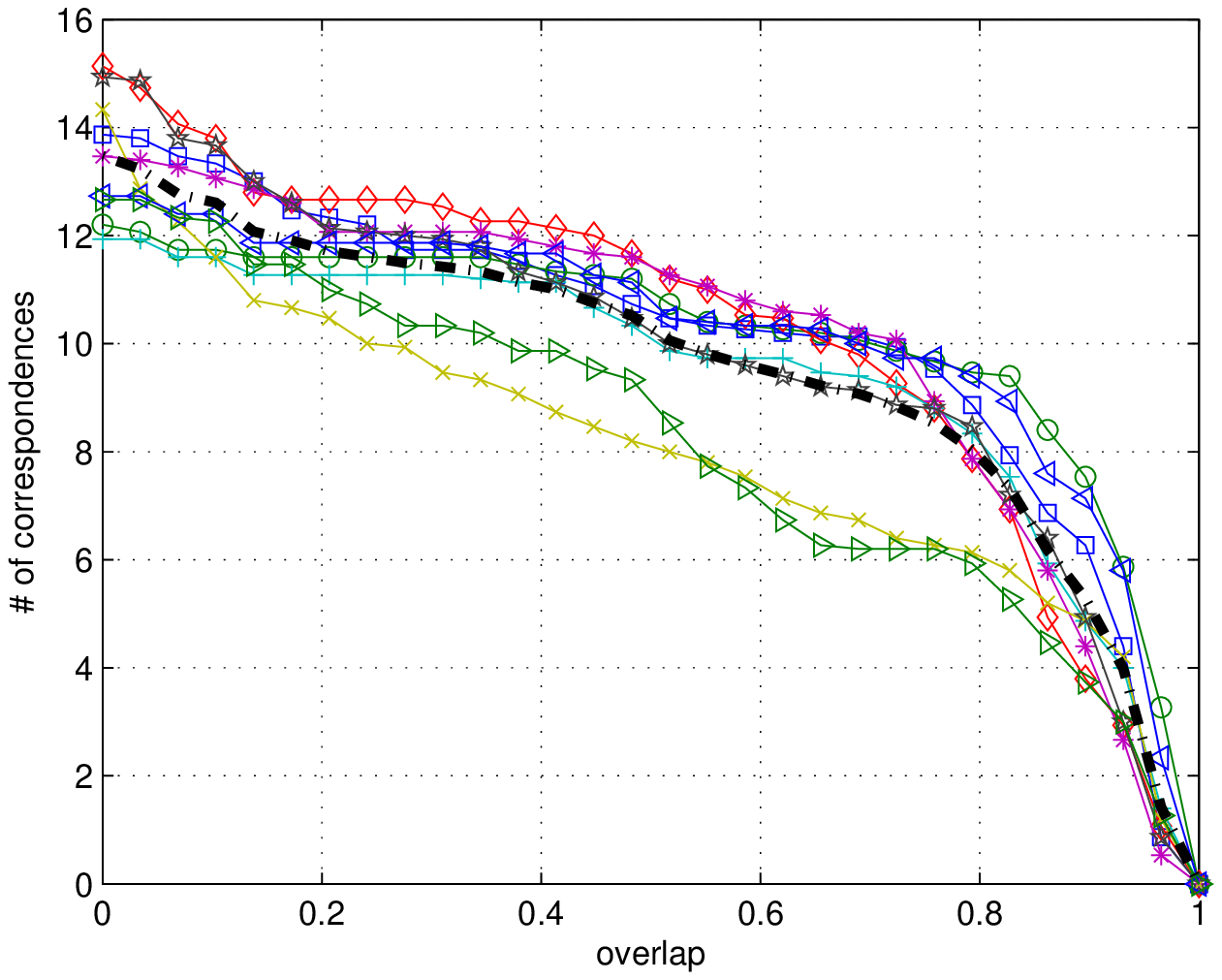} \\
    \end{tabular}
\caption{Repeatability of maximally stable components with the
vertex weight $h_t(v,v)$ (first row) and
edge weight $1/ h_t(v_1,v_2)$ (second row), $t=2048$.}
\label{fig_repeatability}
\end{figure*}

\begin{figure*}[t]
    \includegraphics[width=\linewidth]{PerClass_Legend_H.eps} \\
     \begin{tabular}{cc}
        \includegraphics[width=\columnwidth]{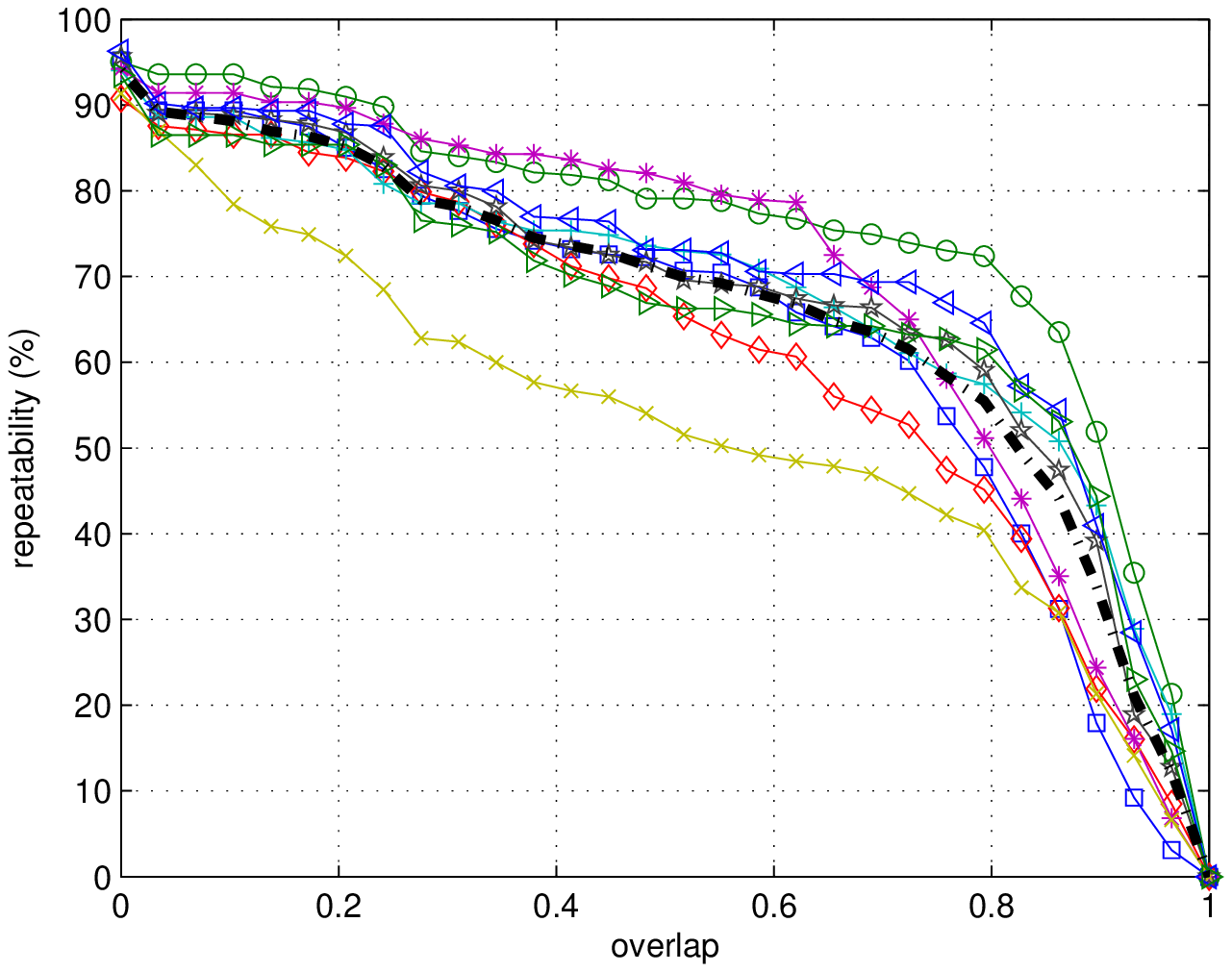} &
        \includegraphics[width=\columnwidth]{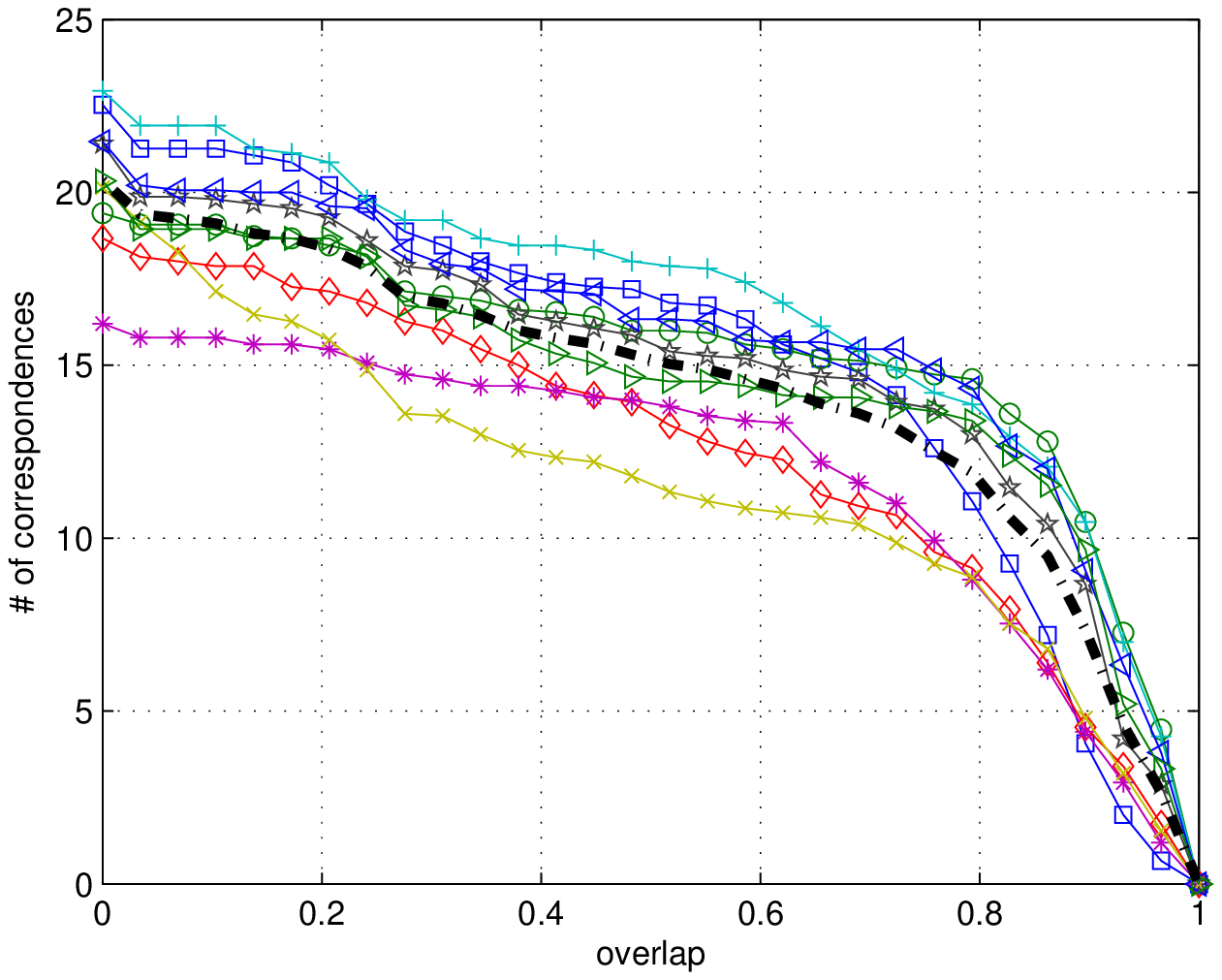} \\

        \includegraphics[width=\columnwidth]{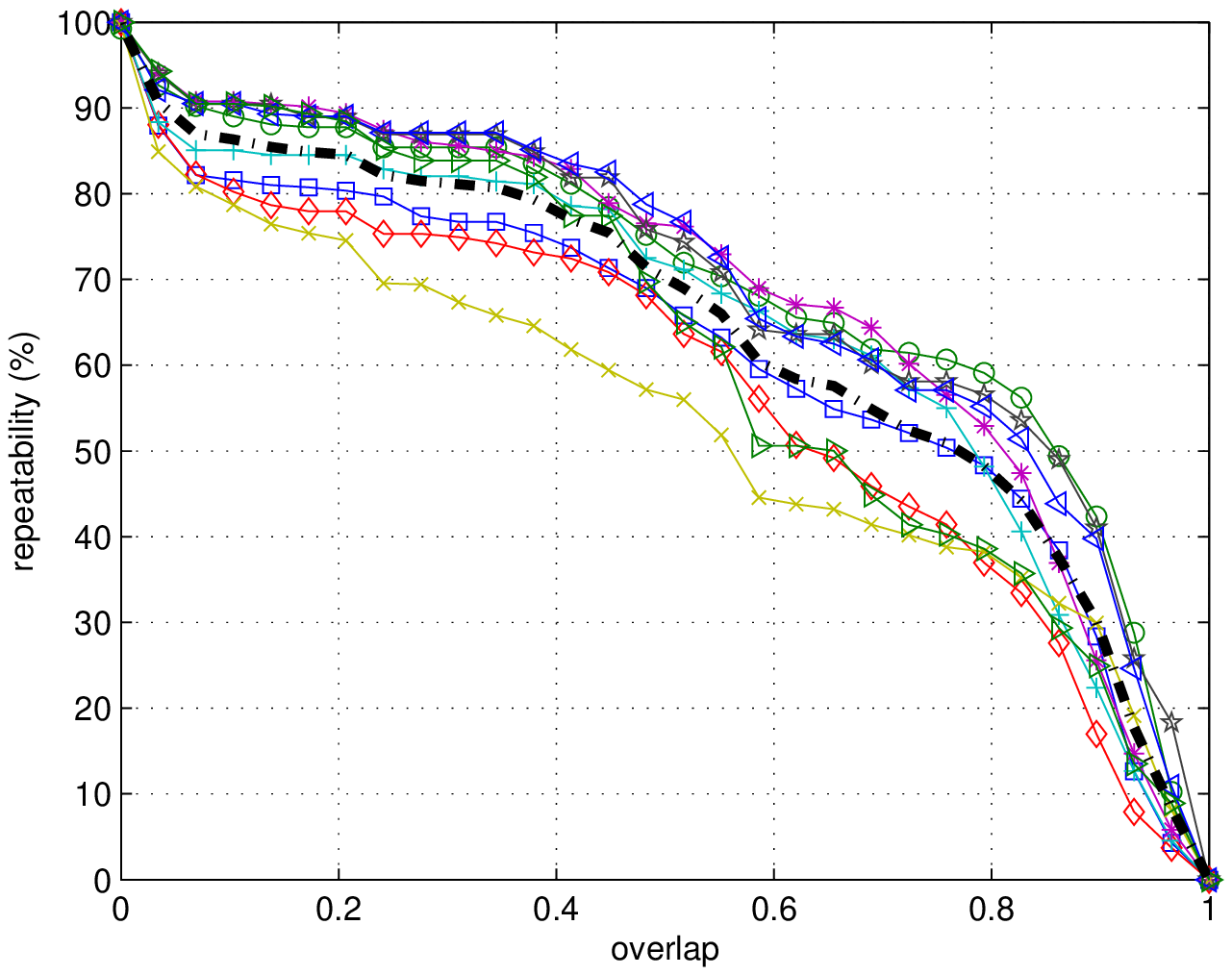} &
        \includegraphics[width=\columnwidth]{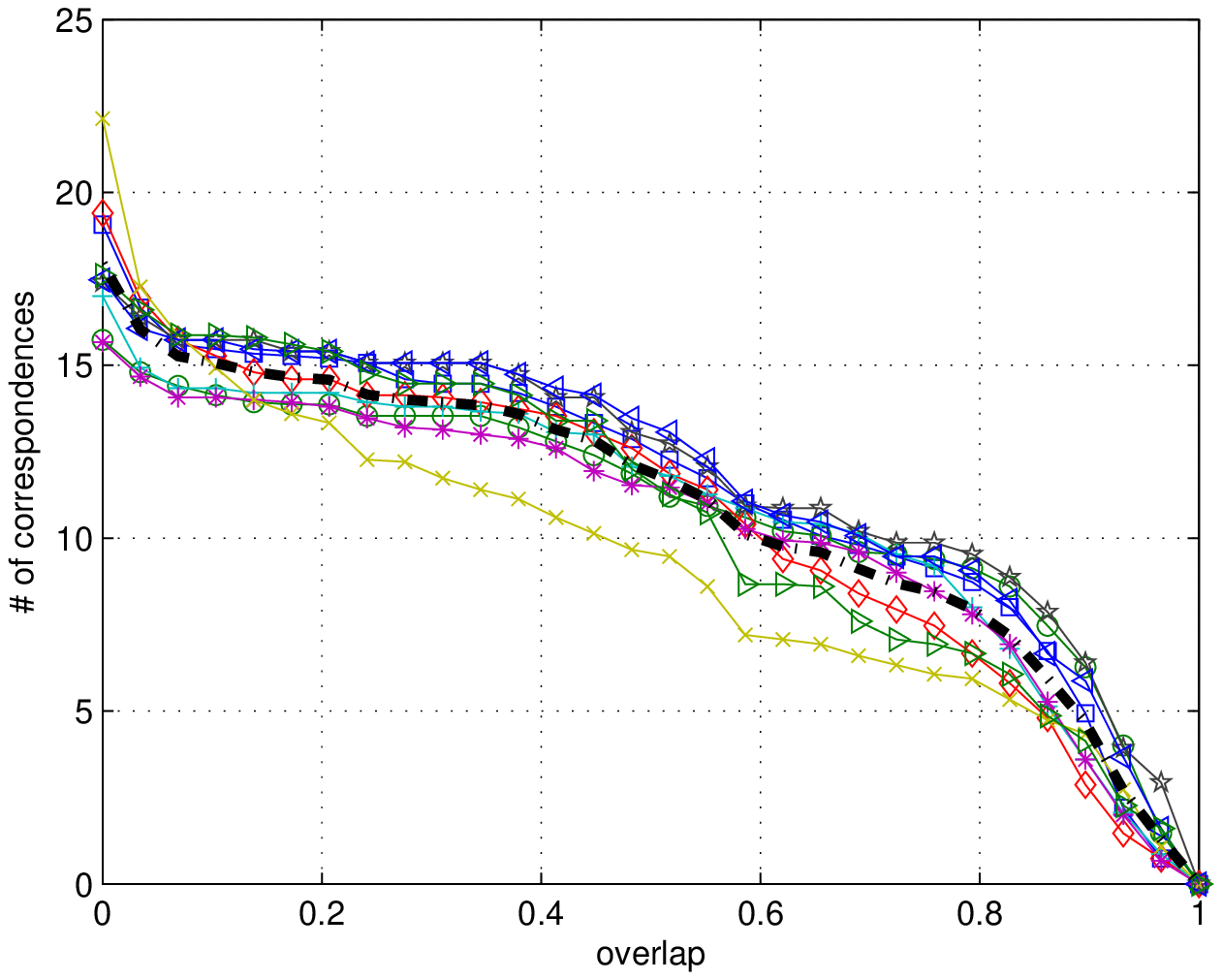} \\
    \end{tabular}
\caption{Repeatability of maximally stable components with the
edge weight $|h_t(v_1,v_1)-h_t(v_2,v_2)|$ (first row) and
edge weight $1/ c(v_1,v_2)$ (second row), $t=2048$.}
\label{fig_repeatability2}
\end{figure*}

%Figures~\ref{fig_region_sample} and \ref{fig_region_parts} show the detected maximally stable regions.
Table~\ref{table_repeatability} summarizes the repeatability of different weighting functions at overlap of $75\%$.
Figures~\ref{fig_repeatability} and \ref{fig_repeatability2} depict the repeatability and the number of correctly matching
regions as the function of the overlap for the best four of the compared weighting functions.
We conclude that scale-dependent weighting generally outperform their scale-invariant counterparts in terms of repeatability.
%This could be explained by the fact that we have selected the best time value for our dataset's common scale, whereas scale-invariant methods suffer from its larger degree of freedom.
The four scalar fields corresponding to different auto-diffusivity functions perform well both when used as vertex and edge weights. Best repeatability is achieved by
the edge weighting function $1/h_t(v_1,v_2)$. Best scale invariant weighting is also the edge weight $1/c(v_1,v_2)$.

\begin{table*}[tb]
    \begin{small}
        \begin{center}
            \begin{tabular}[width=\columnwidth]{lcccc}
                 \hline\hline
                 Weighting     & Maximal instability &  Avg. number of      & Num. of correspondences   & Repeatability  \\
                 function      & score used          &  detected regions    & at overlap$=75\%$         & at overlap$=75\%$ \\ % name used in the code:
                \hline
                $h_t(v,v), t=2048$                                                  & $\infty$& $11.5$ & $6.89$ & $65\%$ \\      %vw+ t=2048
                $c(v,v)$                                                            & $\infty$& $28.3$ & $5.01$ & $18\%$  \\     %vw+ CT
                $\hat{h}_\omega(v,v), \omega = 0$                                   & $\infty$& $15.1$ & $5.43$ & $40\%$ \\      %vw+ siHks at0
                $\|\hat{h}_\omega(v,v)\|_\omega$                                    & $\infty$& $20.2$ & $1.96$ & $10\%$ \\      %vw+ siHksNorm
                 \hline
                $| h_t(v_1,v_1) - h_t(v_2,v_2)|, t=2048$                            & $2.51\times 10^{5}$& $21.7$ & $12.73$ & $60\%$ \\     %ew min scalar HK t=2048
                $| c(v_1,v_1) - c(v_2,v_2) |$                                       & $\infty$  & $11.4$ & $4.12$ & $35\%$ \\      %ew min scalar CT
                $| \hat{h}_\omega(v_1,v_1) - \hat{h}_\omega(v_2,v_2)|, \omega = 0$  & $\infty$  & $71.8$ & $3.33$ & $5\%$ \\       %ew min SIHKS at0
%                $| \|\hat{h}_\omega(v1,v1)\|_\omega - \|\hat{h}_\omega(v1,v1)\|_\omega|$ & $11.5$& $88.5$ & $7.77$ & $9\%$ \\ %ew min siHksNorm
                $1/h_t(v_1,v_2), t=2048$                                            & $\infty$  & $13.8$ & $8.64$ & $68\%$ \\      %ew min inverse HKS t=2048
                $1/c(v_1,v_2)$                                                      & $1$    & $11.1$ & $5.07$ & $45\%$ \\      %ew min inverse CT
                $1/\|\hat{h}_\omega(v1,v2)\|_\omega$                                & $100$    & $4.8$ & $1.81$ & $45\%$ \\       %ew min inverse SIHKS
%                $1/\|\hat{h}_\omega(v,v)\|_\omega (altered)$ & $11.5$& $67.5$ & $7.68$ & $12\%$ \\           %ew min inverse SIHKS diff
                $\| h_t(v_1,\cdot) - h_t(v_2,\cdot)\|_X, t=2048$                    & $5\times 10^{6}$& $13.8$ & $7.19$ & $50\%$ \\      %ew min diffusion distance t=2048
                $\| h_t(v_1,v_1) - h_t(v_2,v_2) \|_t, t\in[128,32\times 10^3]$      & $1.58\times  10^{7}$& $18.4$ & $9.99$ & $58\%$ \\    %ew min time_integral
                 \hline\hline
            \end{tabular}
        \end{center}
    \end{small}
    \caption{Repeatability of maximally stable components with different vertex and edge weighting functions.}
    \label{table_repeatability}
\end{table*}

\subsection{Descriptor discriminativity}

\begin{figure*}[t]
\begin{small}
    \begin{center}
            \includegraphics[width=\columnwidth]{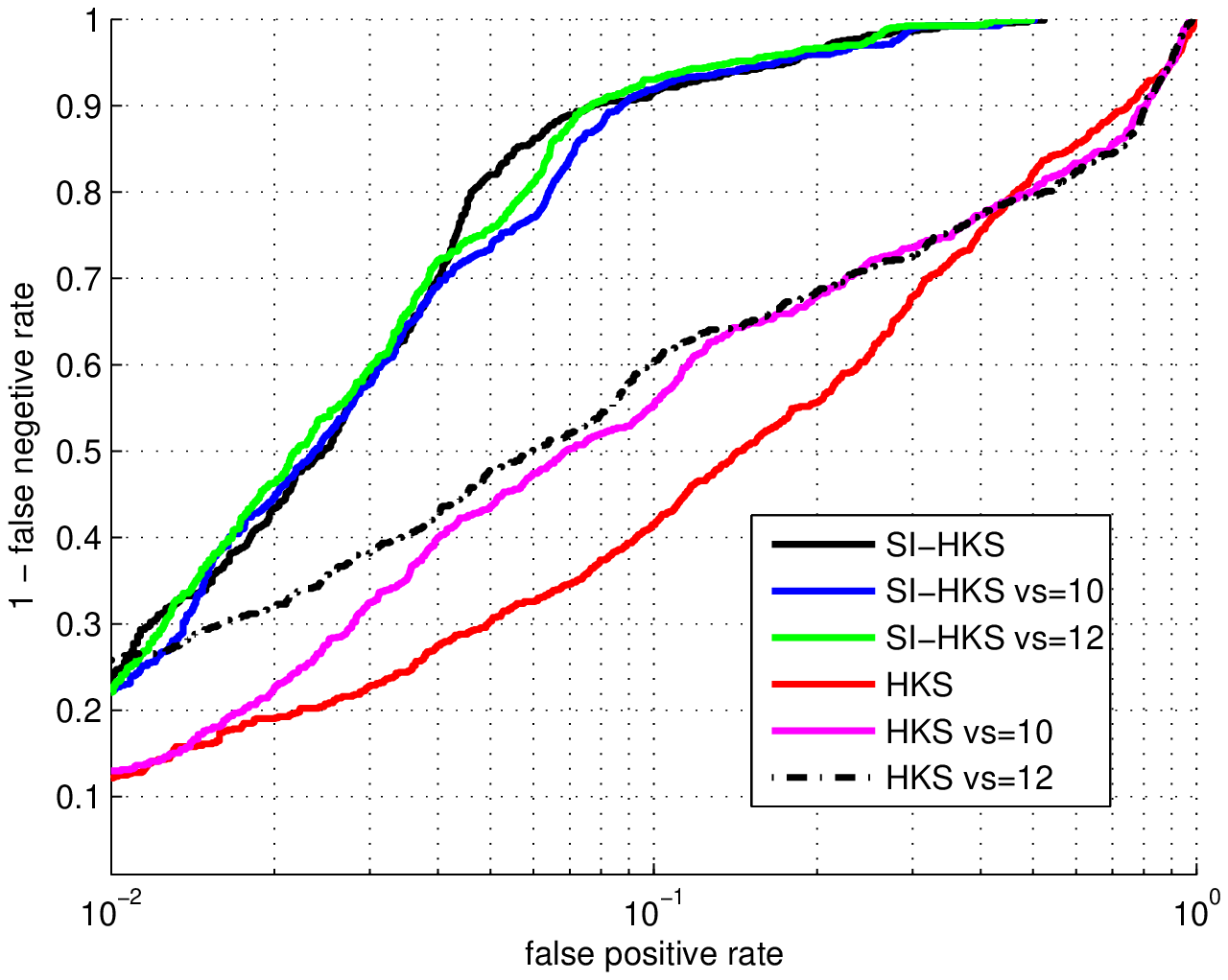}
            \includegraphics[width=\columnwidth]{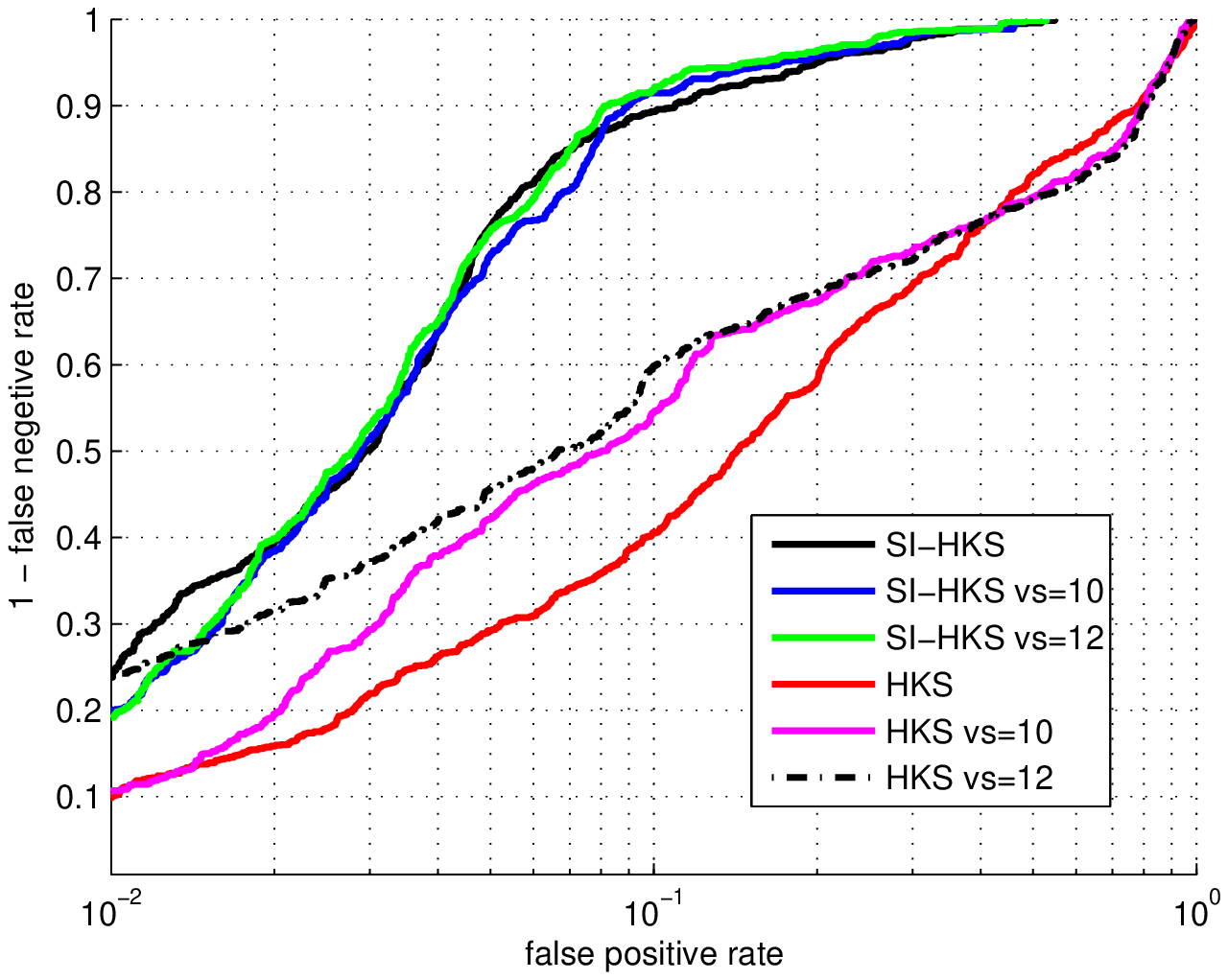}\\
\ifx\isfulltr\undefined
\else
            \includegraphics[width=\columnwidth]{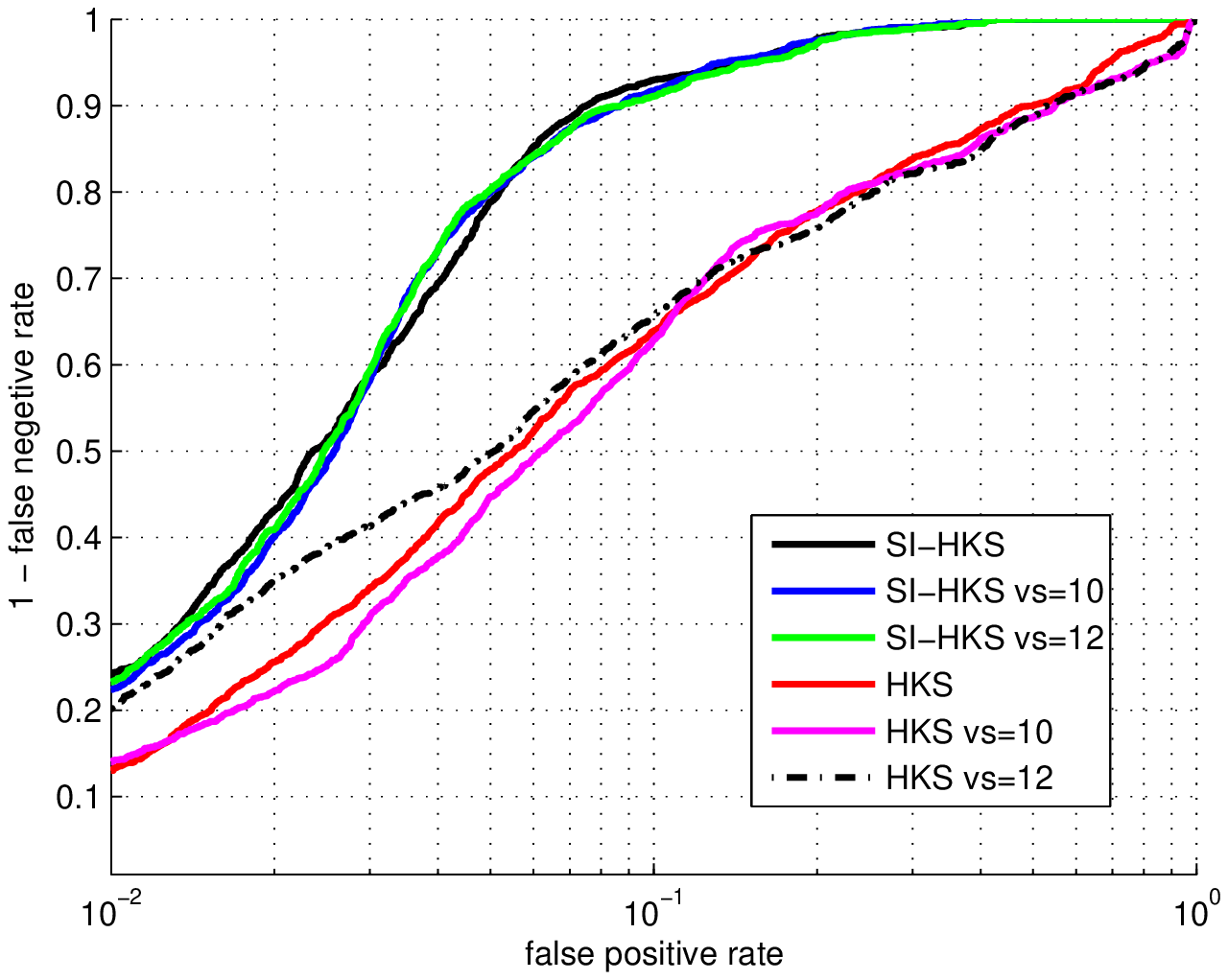}
            \includegraphics[width=\columnwidth]{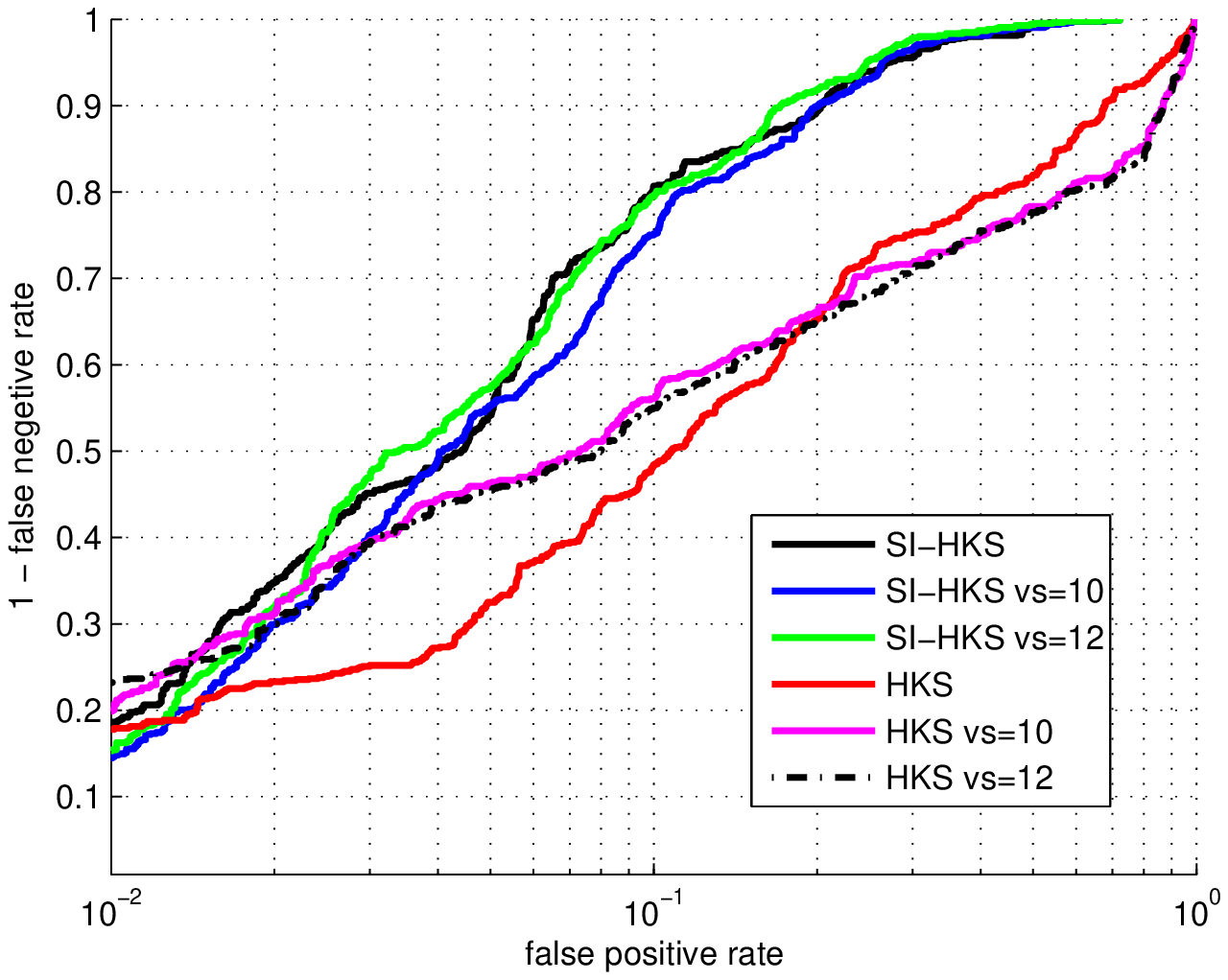}\\
\fi
    \end{center}
\caption{
    ROC curves of different regions descriptors (``vs'' stands for vocabulary size).
\ifx\isfulltr\undefined
    The following detectors were used: vertex weight $h_t(v,v)$ (left), and edge weight $1/h_t(v_1,v_2)$ (right).
\else
    The following detectors were used (left-to-right, top-to-bottom):
    vertex weight $h_t(v,v)$, edge weight $1/h_t(v_1,v_2)$,
    edge weight $| h_t(v_1,v_1) - h_t(v_2,v_2)|$, and edge weight $1/c(v_1,v_2)$.
\fi
\label{fig_ROCs}
    }
\end{small}

\end{figure*}

\begin{figure*}[tb]
\begin{small}
    \begin{center}
    \includegraphics[width=.9\linewidth]{PerClass_Legend_H.eps} \\
        \includegraphics[width=.5\linewidth]{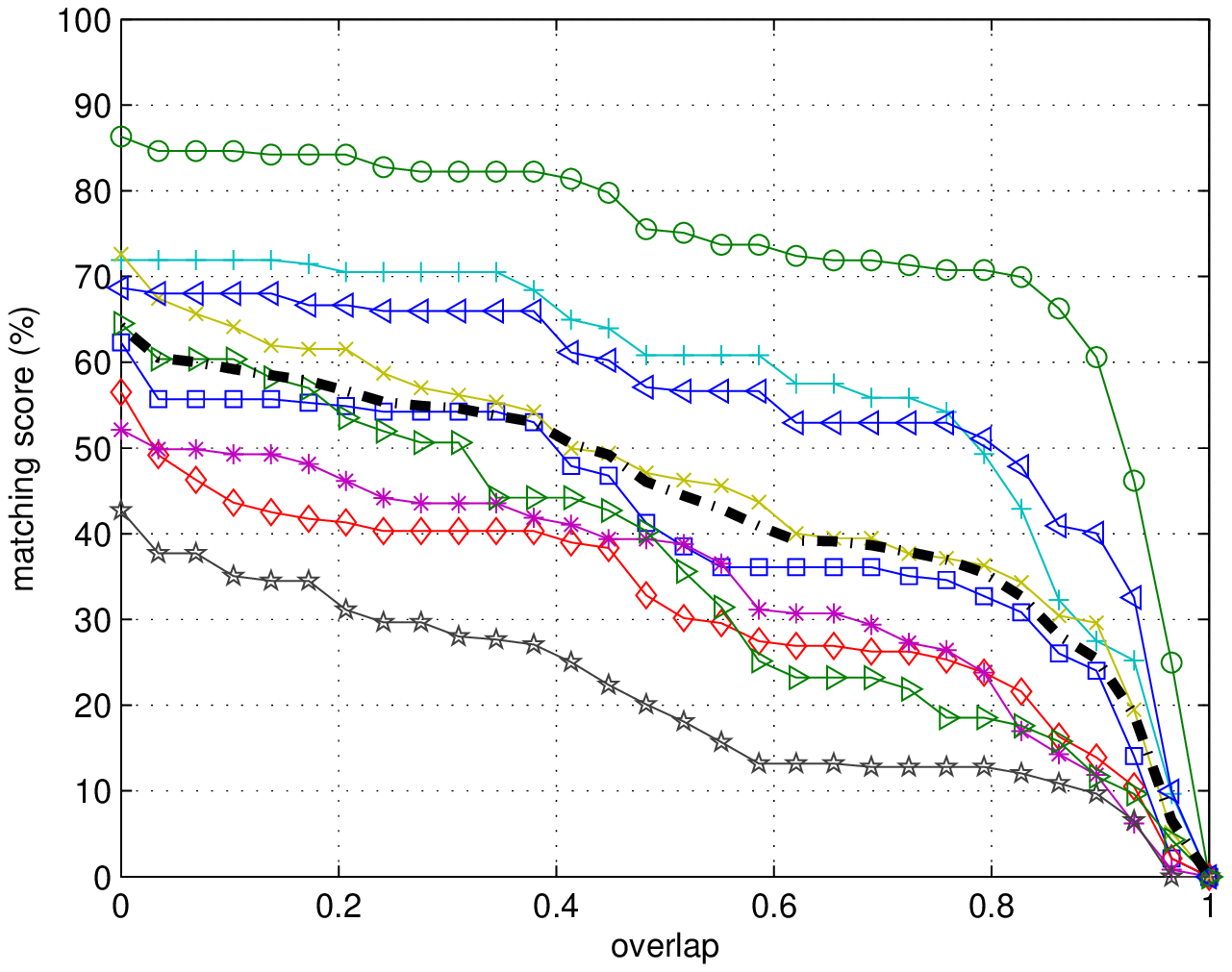} \hspace{-5mm}
        \includegraphics[width=.5\linewidth]{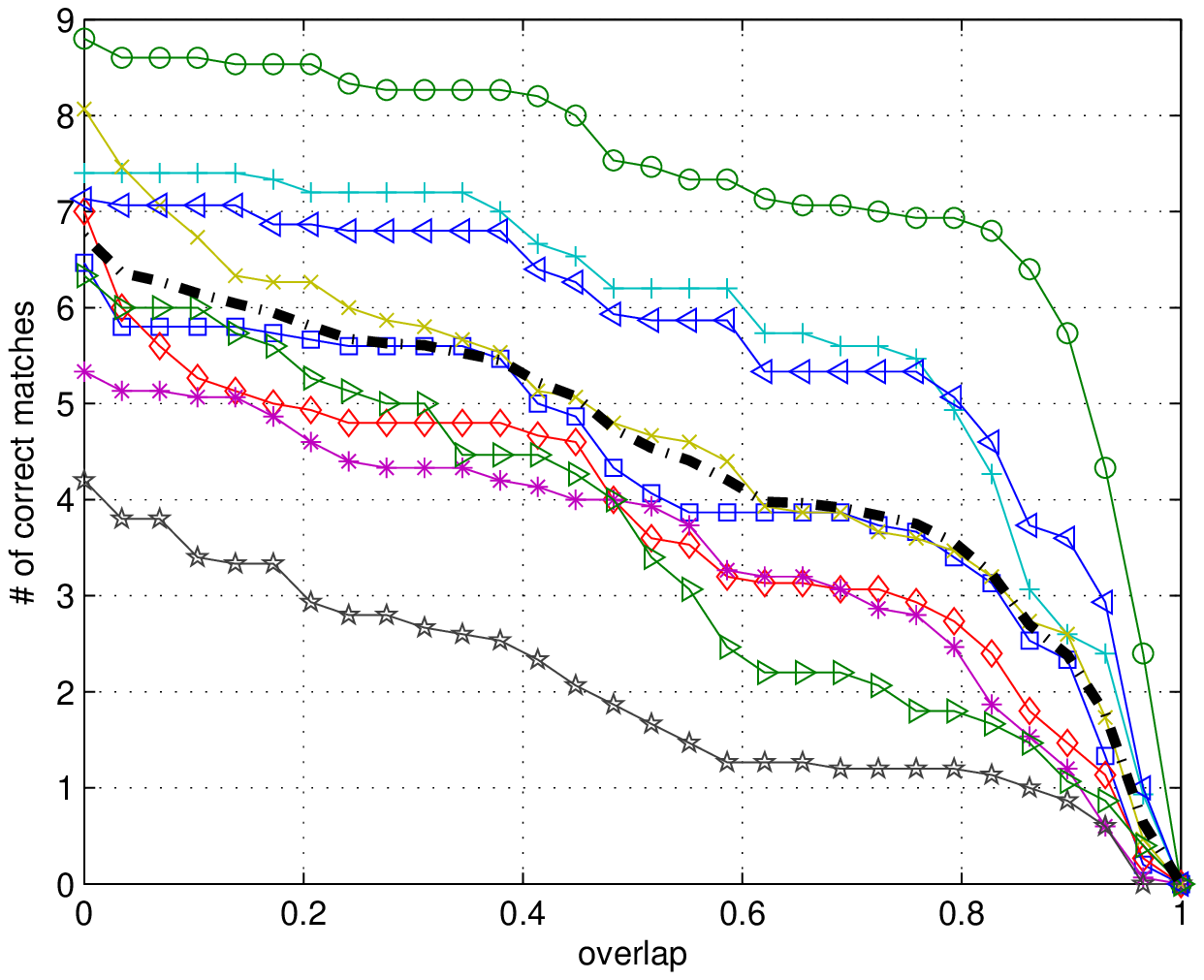} \\
        \includegraphics[width=.5\linewidth]{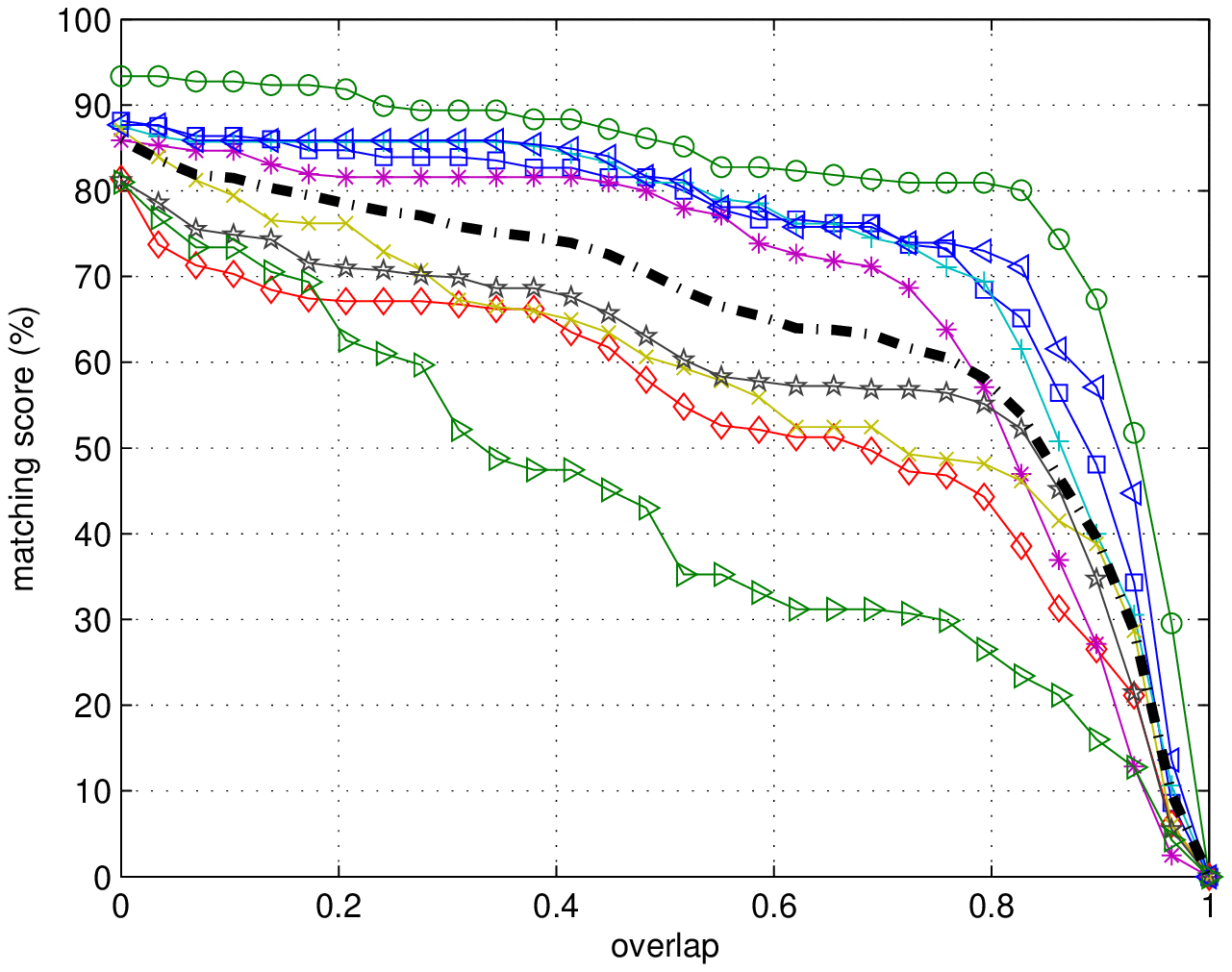} \hspace{-5mm}
        \includegraphics[width=.5\linewidth]{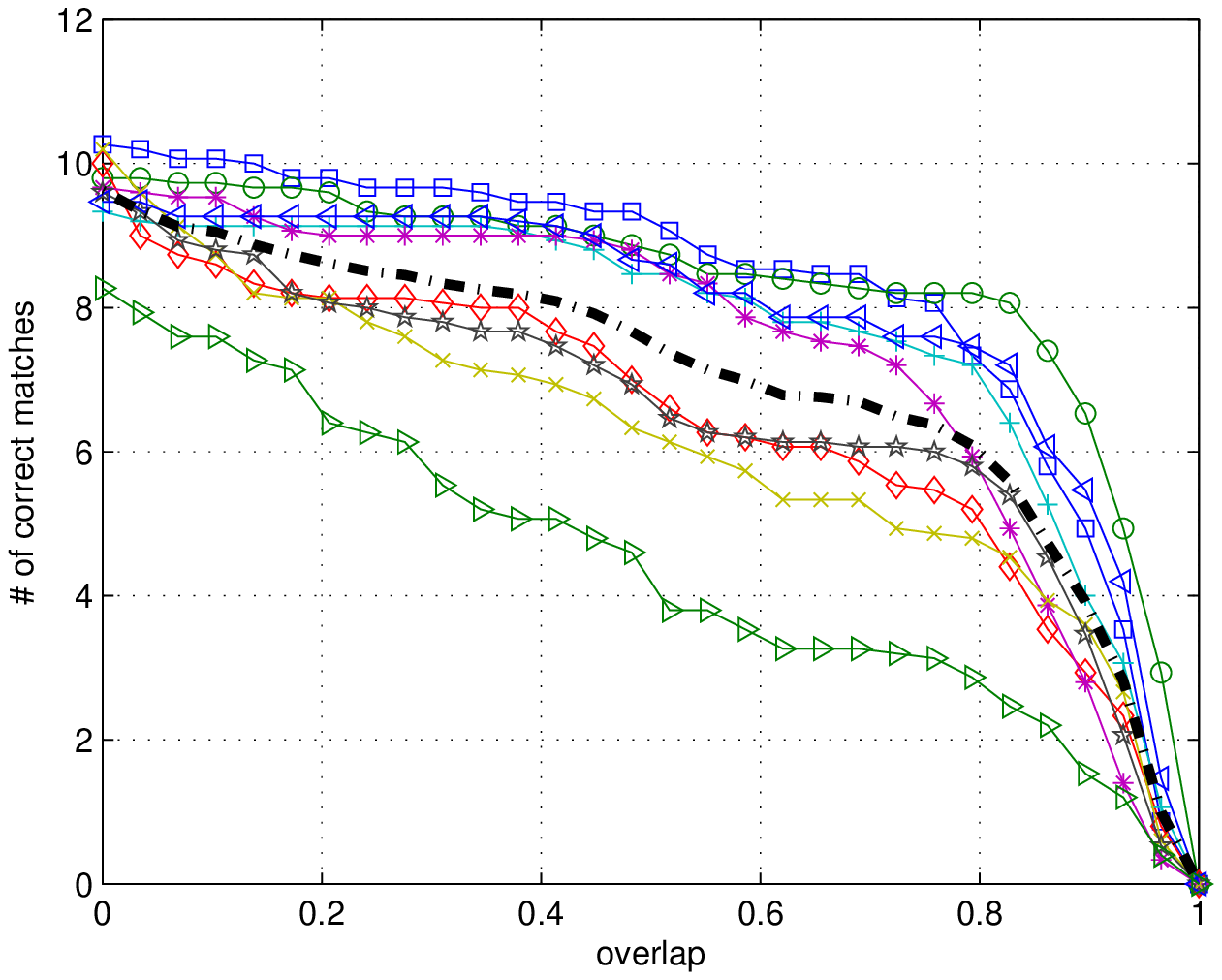} \\
\caption{Performance of region descriptors with regions detected using the vertex weight $h_t(v,v)$, $t=2048$.
Shown are the HKS descriptor (first row) and SI-HKS descriptor (second row). \label{fig_desc_vw} }
    \end{center}
\end{small}
\end{figure*}

% -----------------------------------------------------------------------------------
\begin{figure*}[tb]
\begin{small}
    \begin{center}
    \includegraphics[width=.9\linewidth]{PerClass_Legend_H.eps} \\
        \includegraphics[width=.5\linewidth]{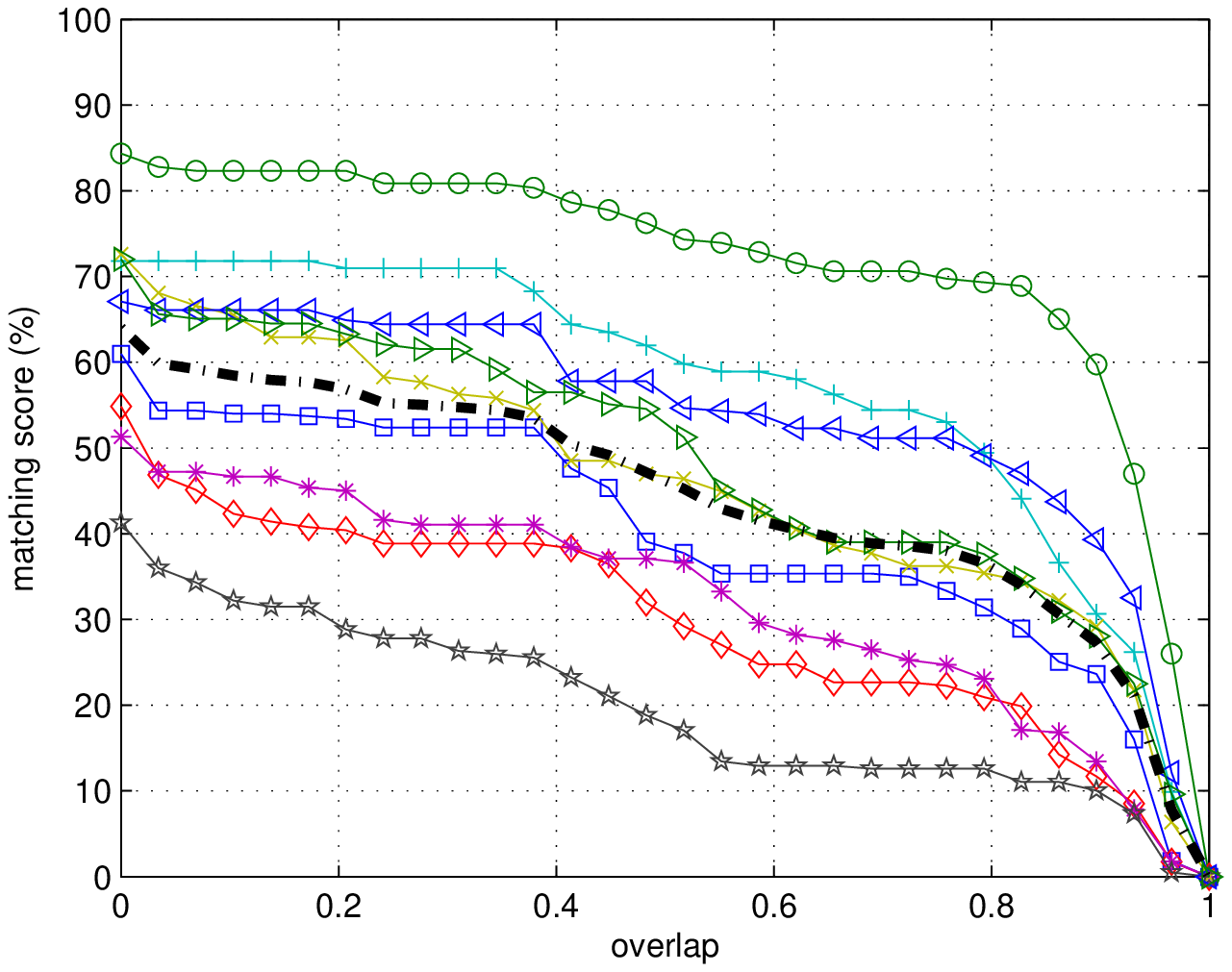} \hspace{-5mm}
        \includegraphics[width=.5\linewidth]{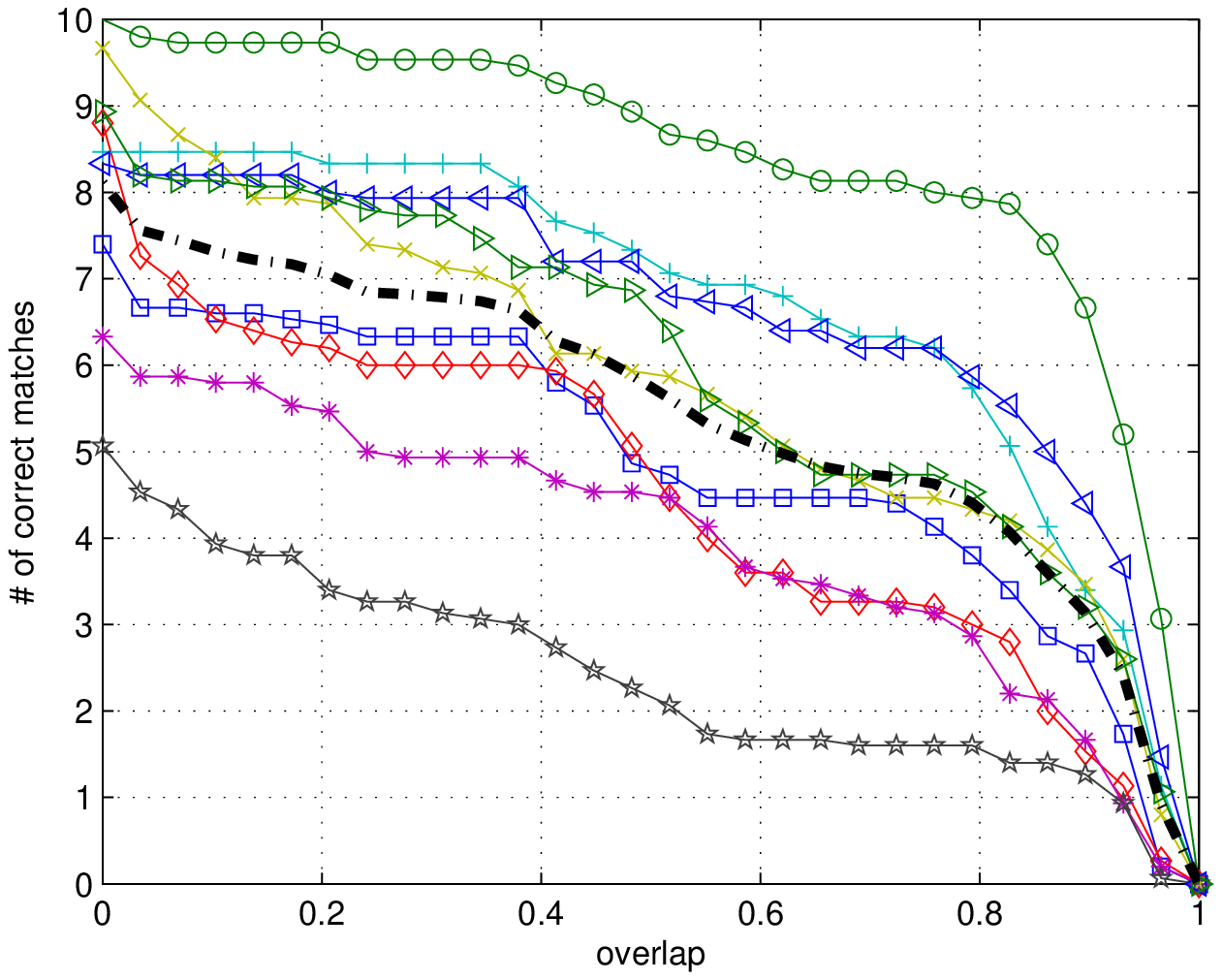} \\
        \includegraphics[width=.5\linewidth]{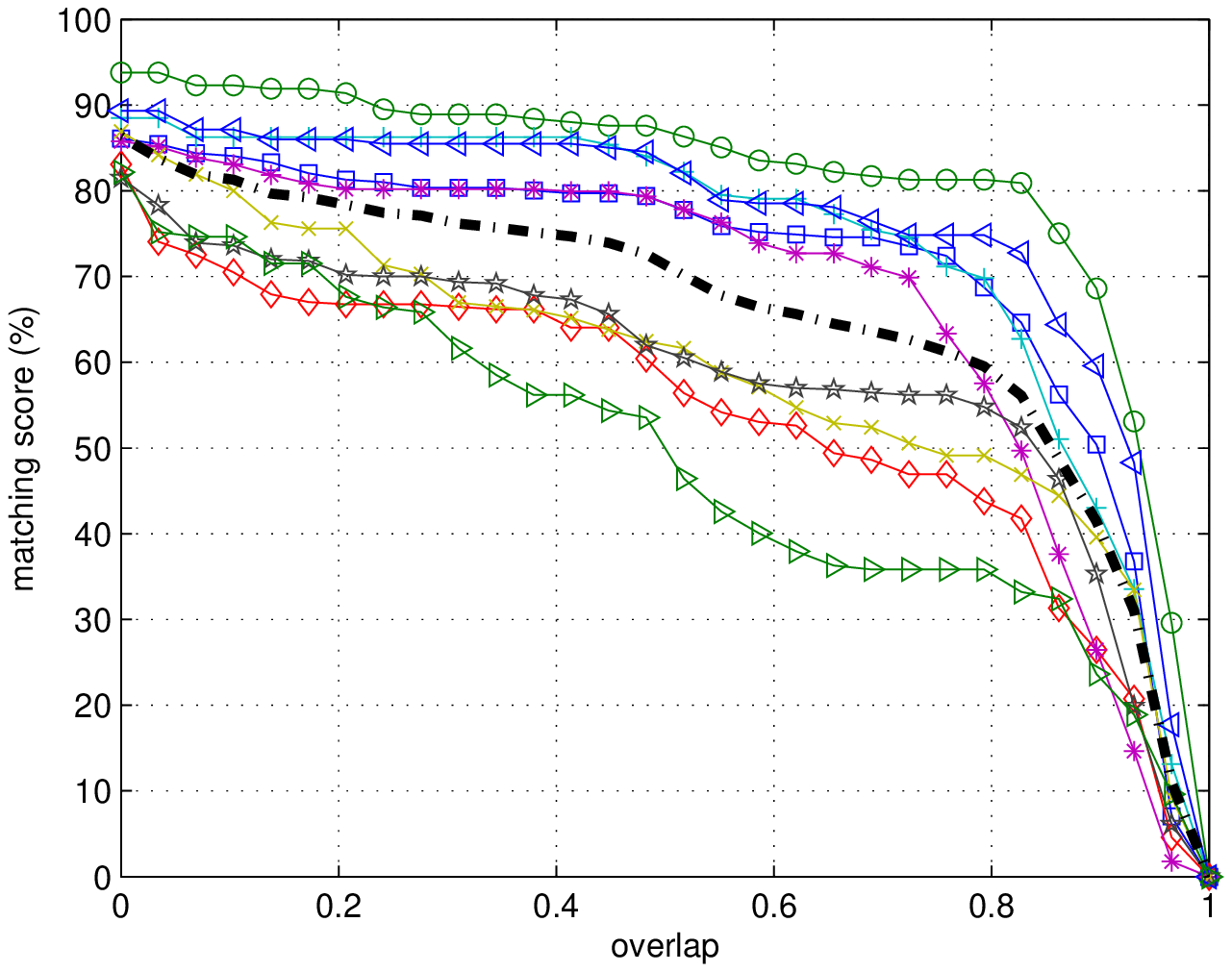} \hspace{-5mm}
        \includegraphics[width=.5\linewidth]{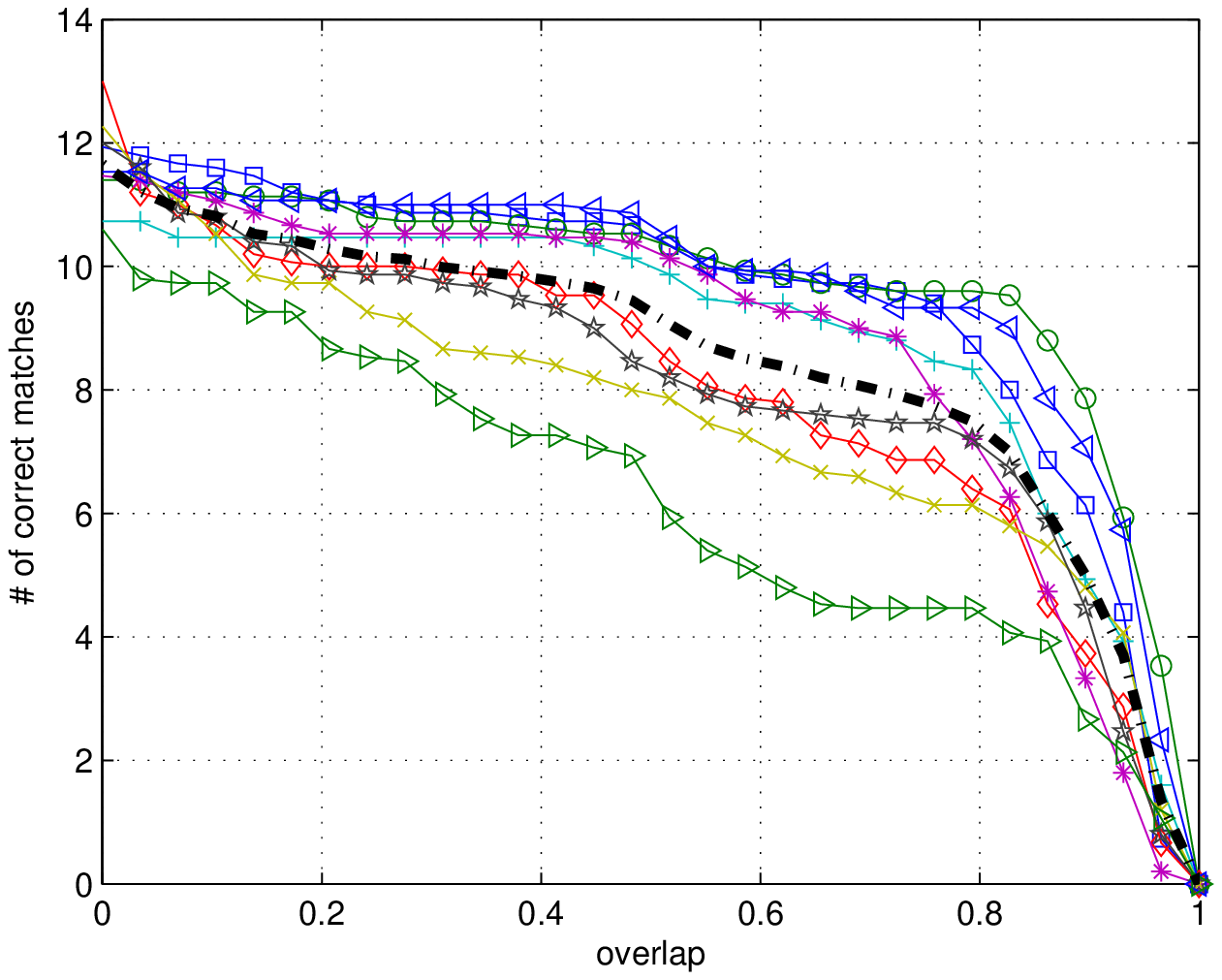} \\
\caption{Performance of region descriptors with regions detected using the  edge weight $1/ h_t(v_1,v_2)$, $t=2048$.
Shown are the HKS descriptor (first row) and SI-HKS descriptor (second row). \label{fig_desc_ew1} }
    \end{center}
\end{small}
\end{figure*}

\ifx\isfulltr\undefined
\else
\begin{figure*}[t]
\begin{small}
    \begin{center}
    \includegraphics[width=.9\linewidth]{PerClass_Legend_H.eps} \\
        \includegraphics[width=.5\linewidth]{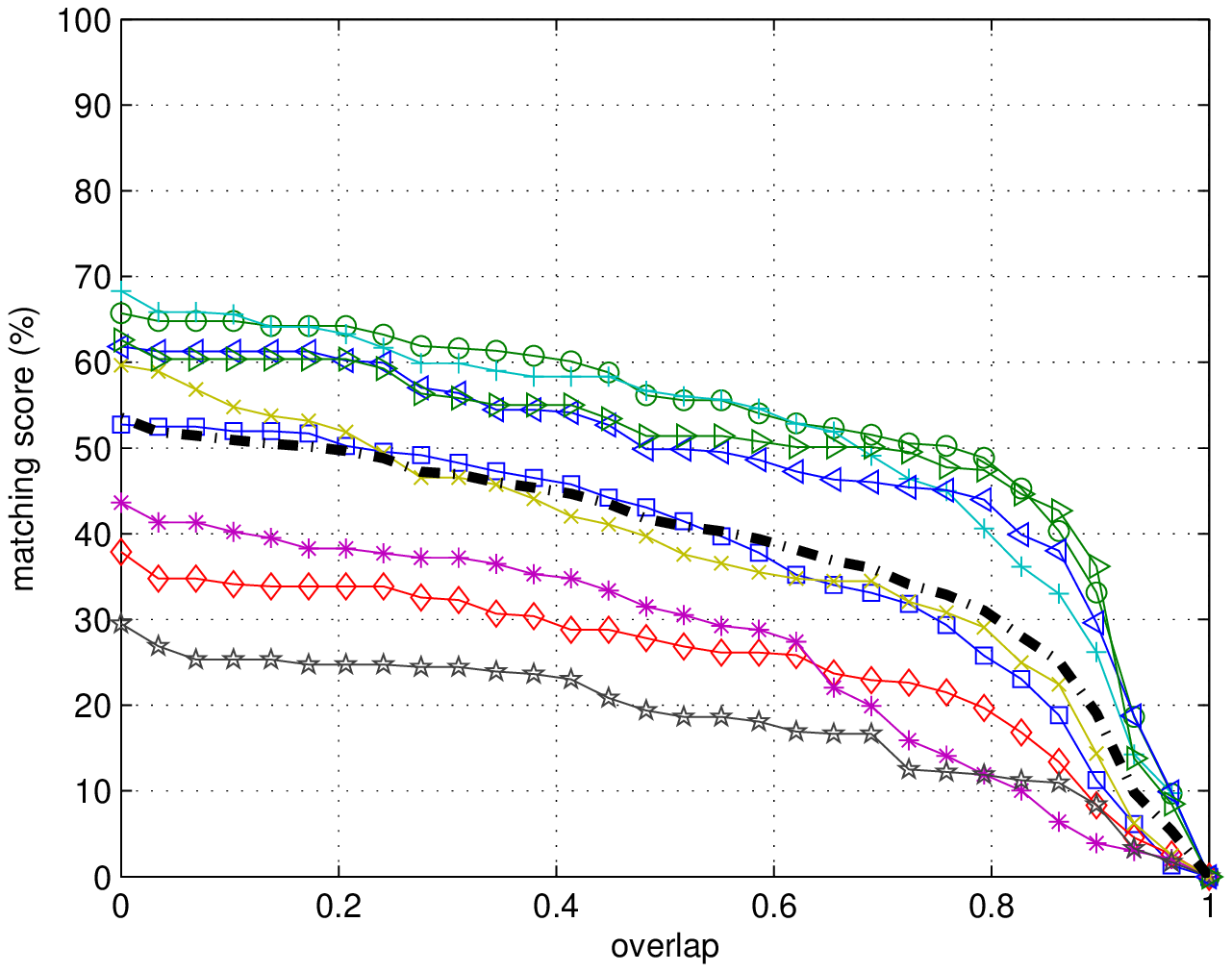} \hspace{-5mm}
        \includegraphics[width=.5\linewidth]{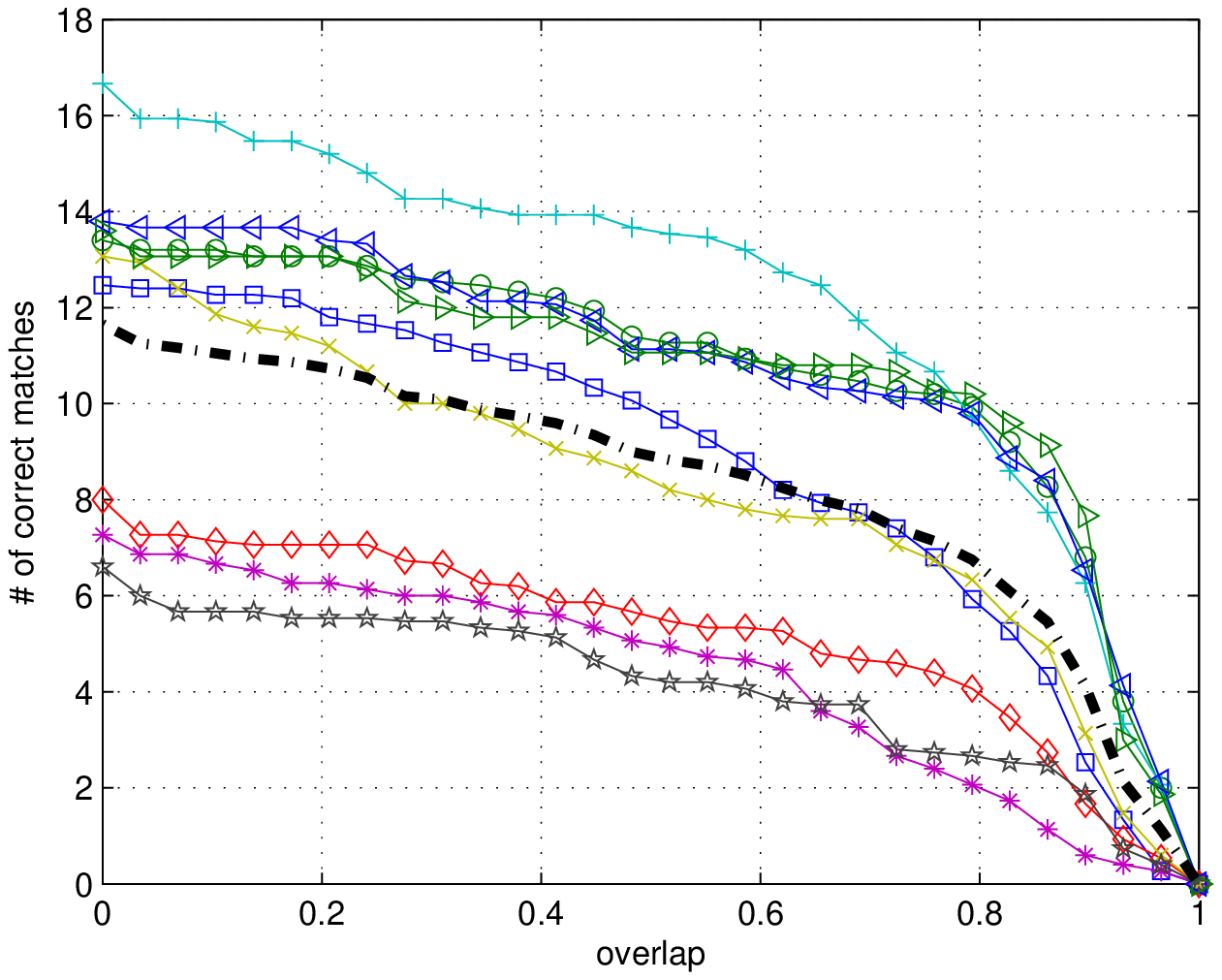} \\
        \includegraphics[width=.5\linewidth]{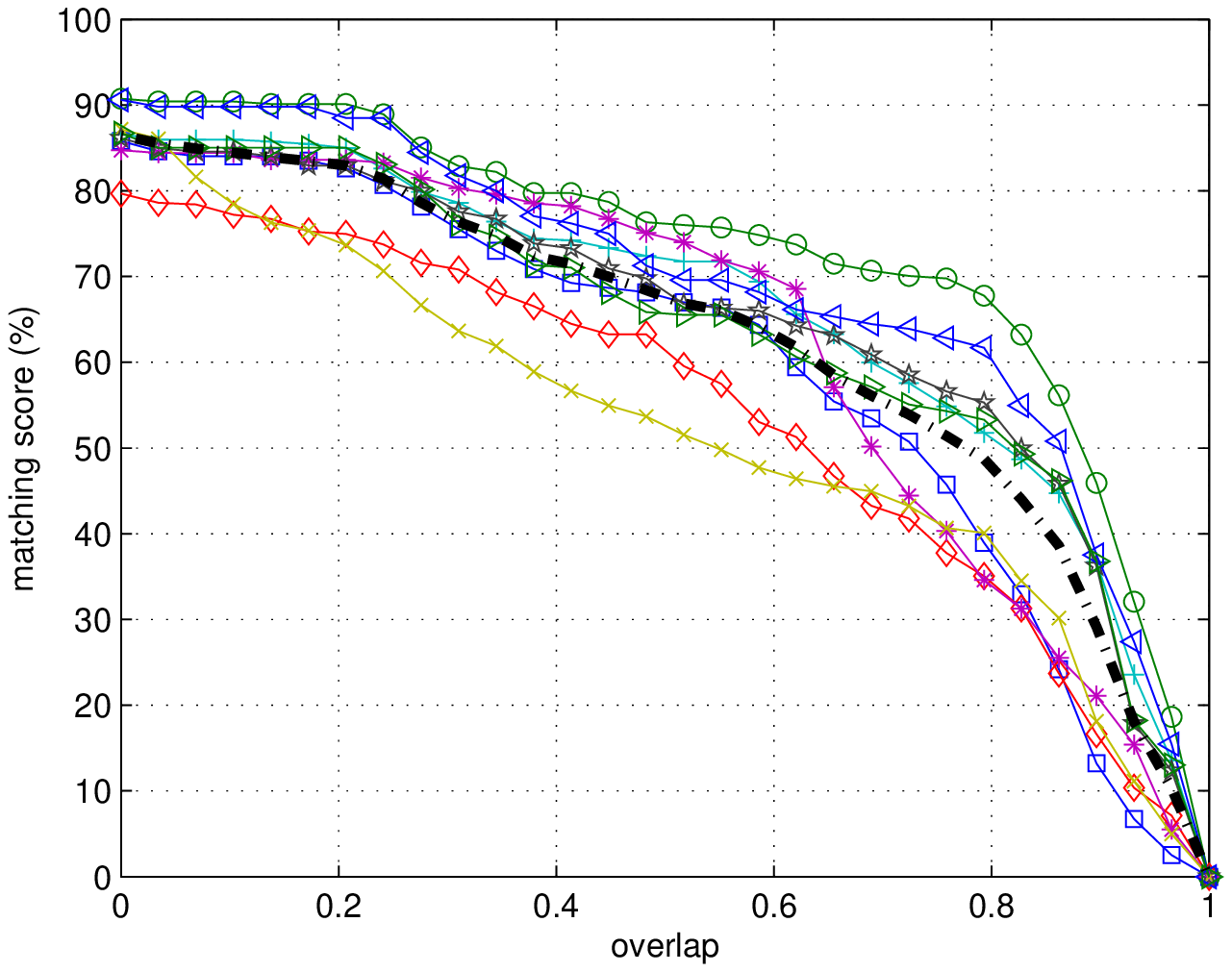} \hspace{-5mm}
        \includegraphics[width=.5\linewidth]{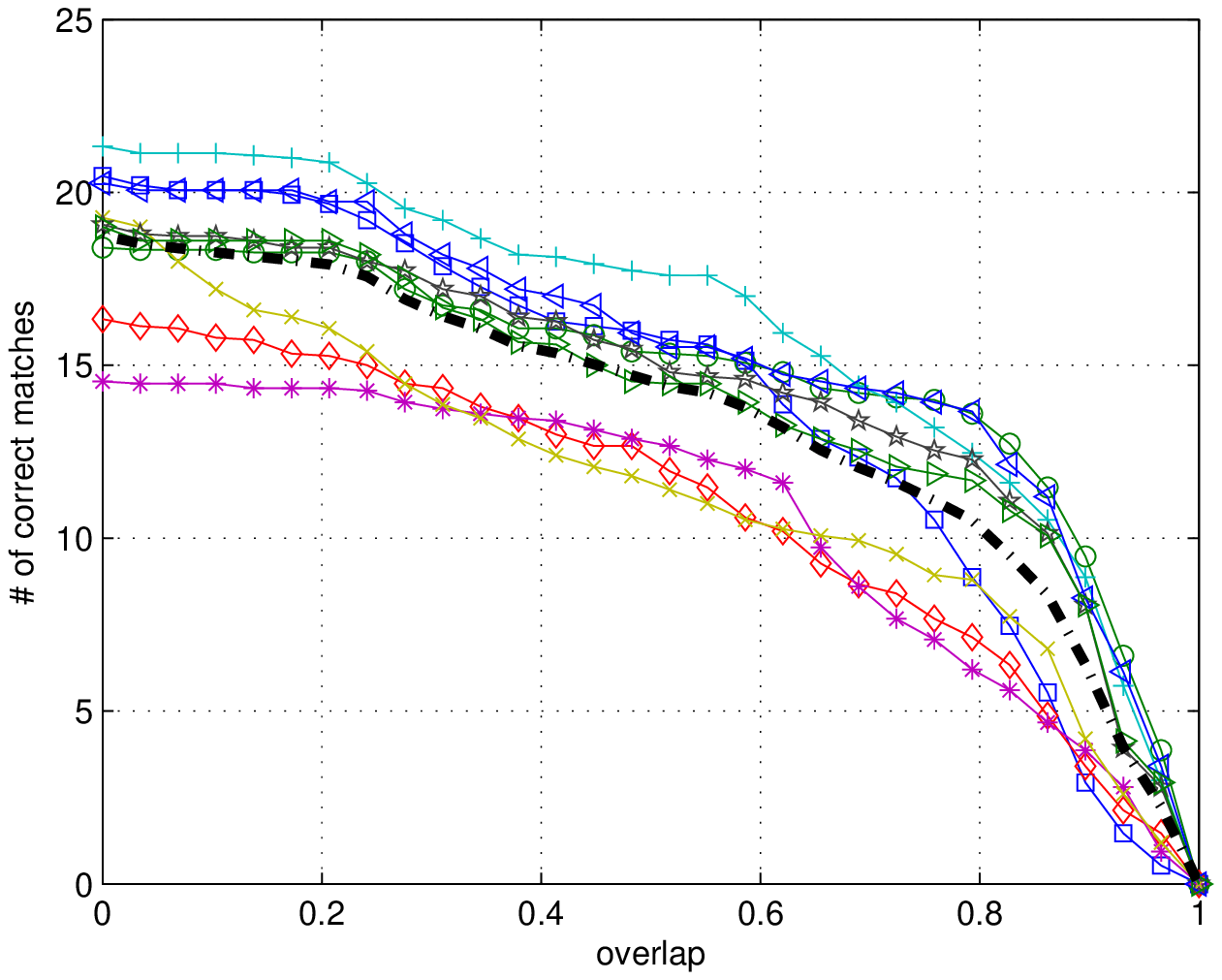} \\
\caption{Performance of region descriptors with regions detected using the  edge weight $| h_t(v_1,v_1) - h_t(v_2,v_2)|$, $t=2048$.
Shown are the HKS descriptor (first row) and SI-HKS descriptor (second row).
\label{fig_desc_ew2} }
    \end{center}
\end{small}
\end{figure*}

\begin{figure*}[t]
\begin{small}
    \begin{center}
    \includegraphics[width=.9\linewidth]{PerClass_Legend_H.eps} \\
        \includegraphics[width=.5\linewidth]{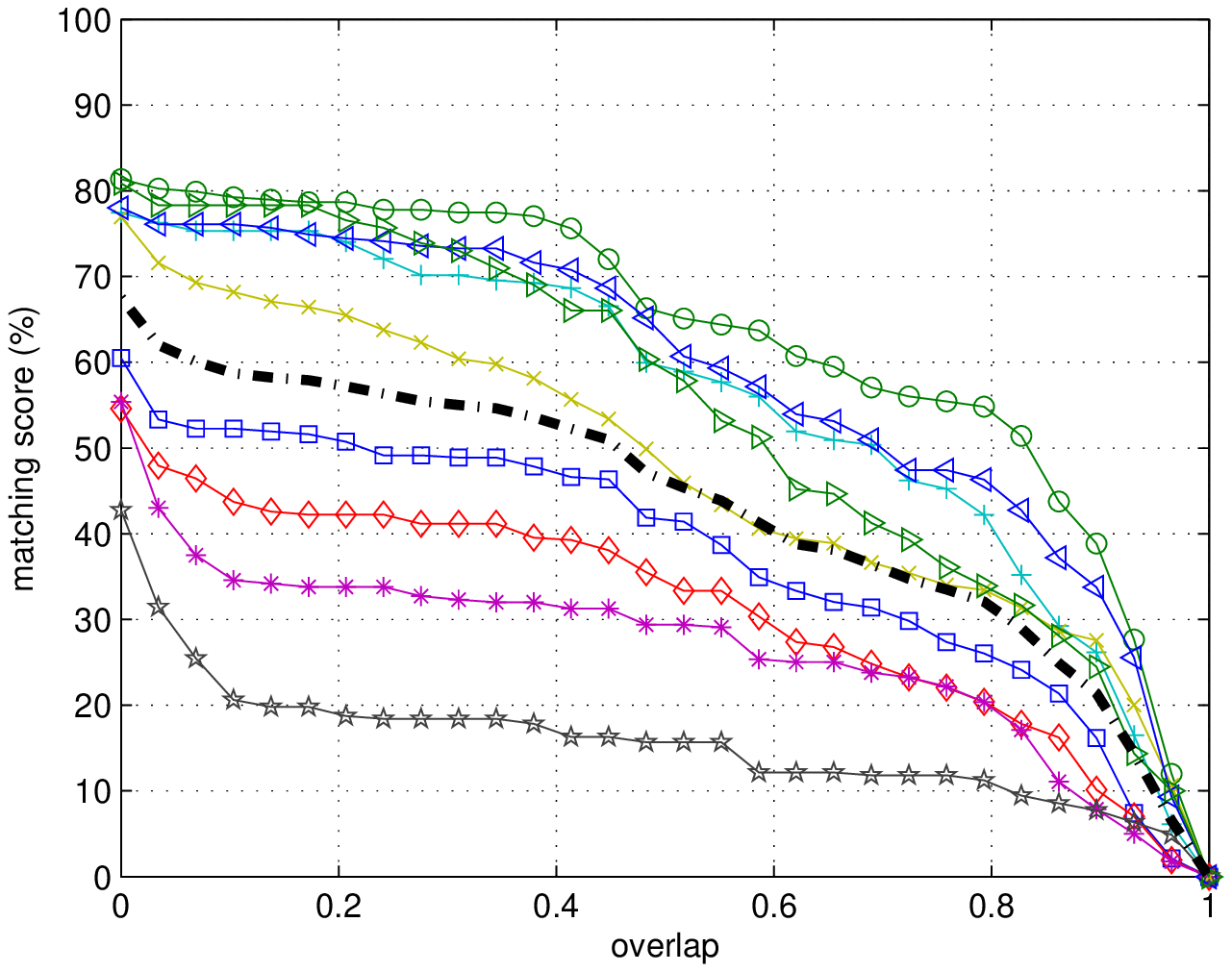} \hspace{-5mm}
        \includegraphics[width=.5\linewidth]{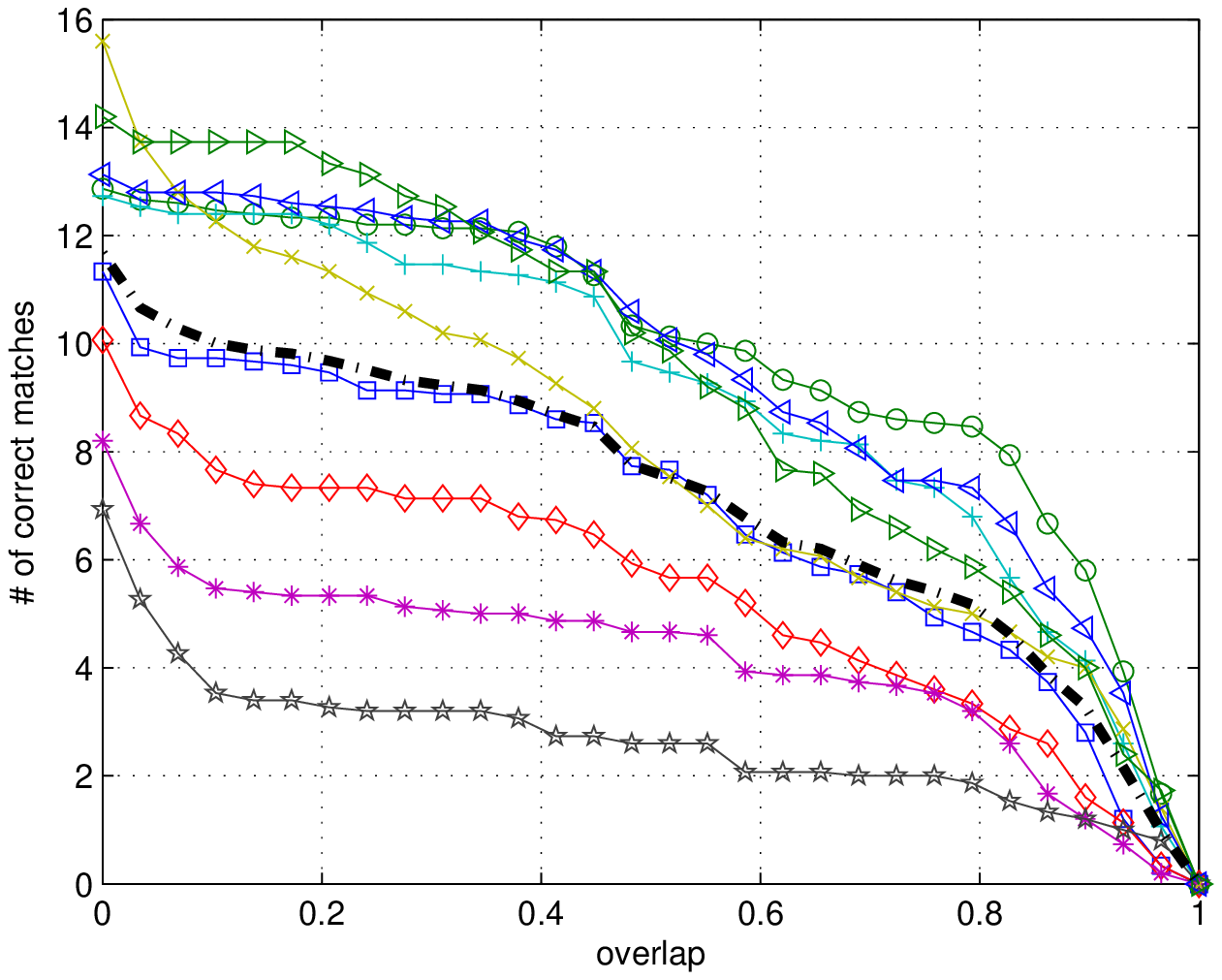} \\
        \includegraphics[width=.5\linewidth]{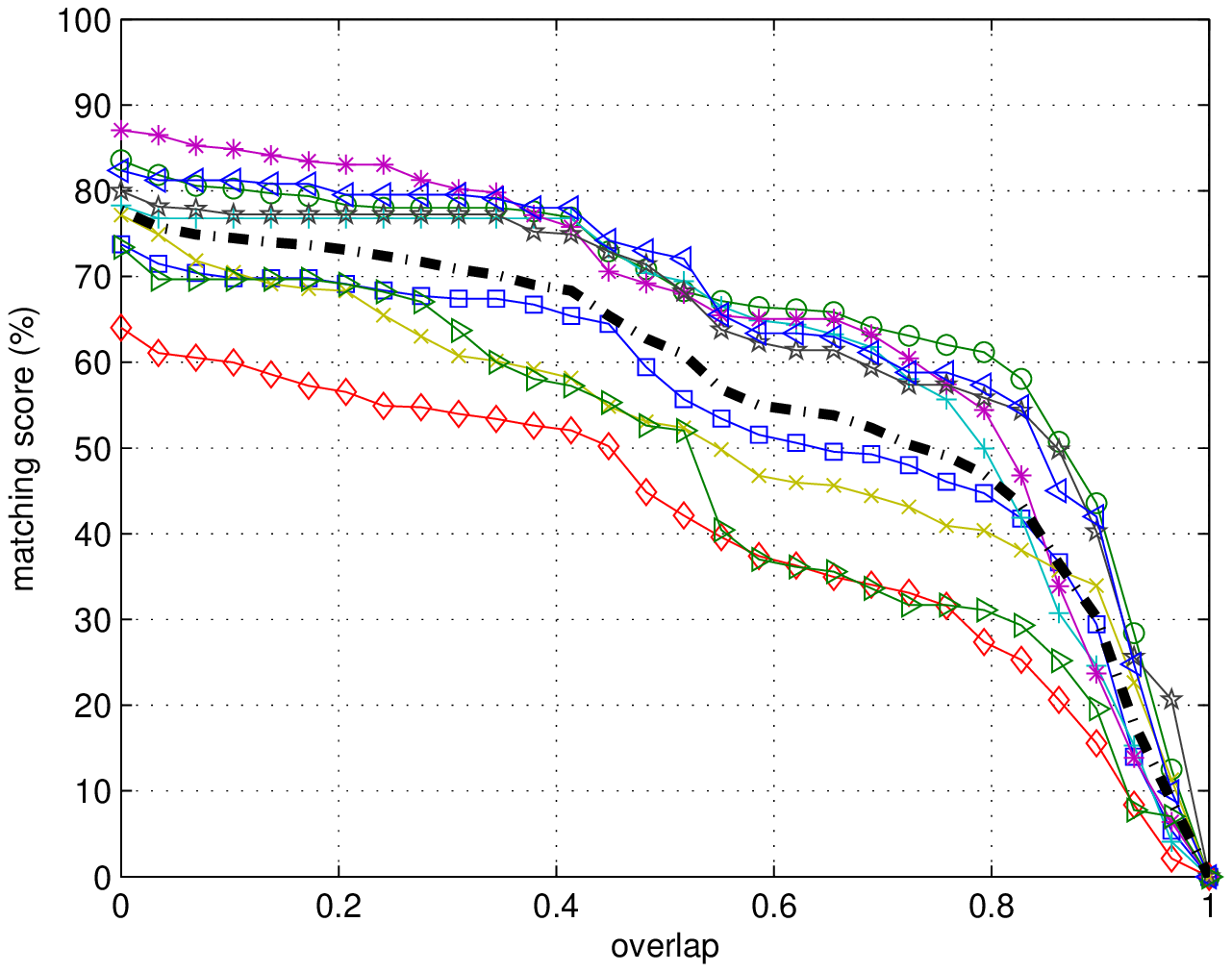} \hspace{-5mm}
        \includegraphics[width=.5\linewidth]{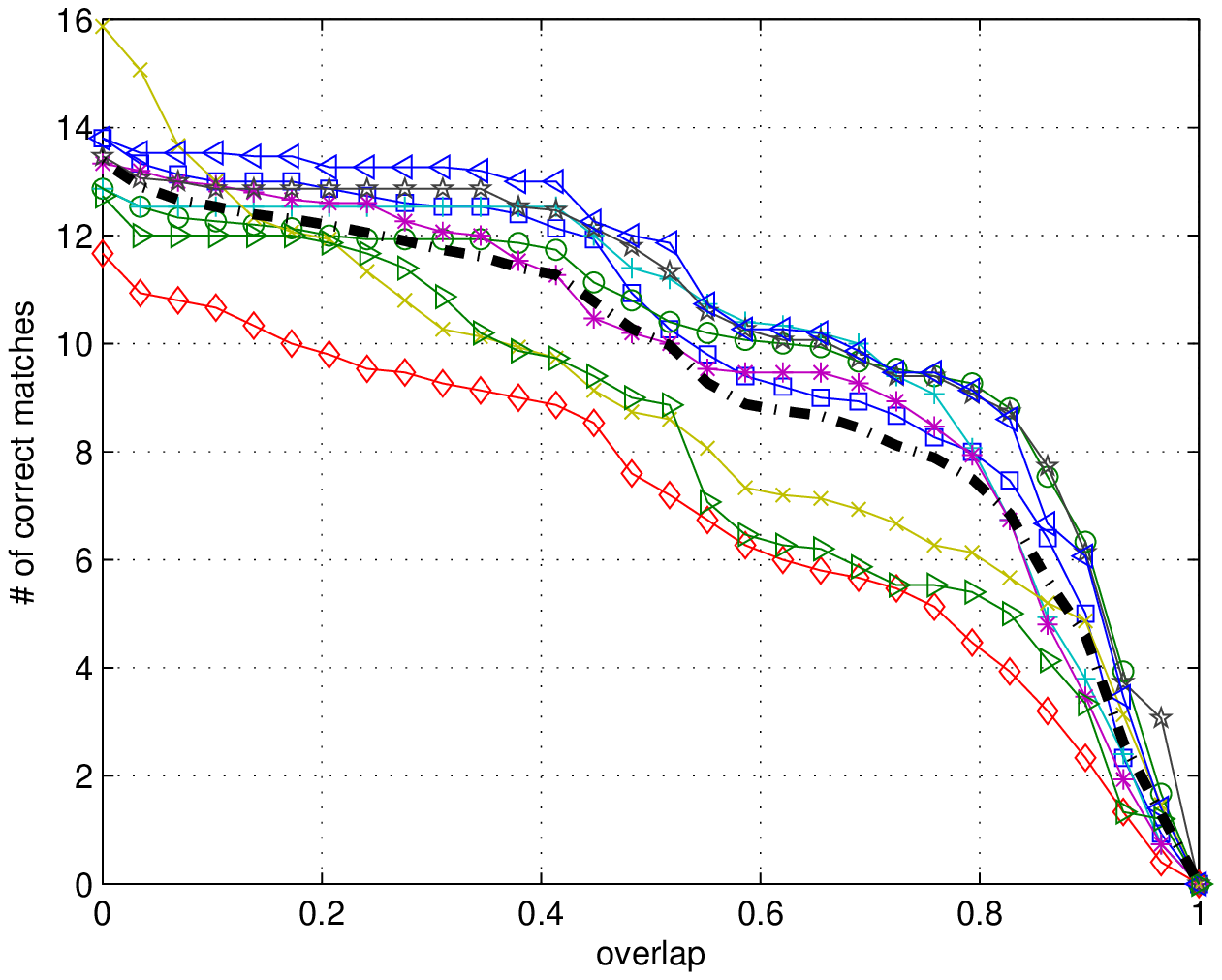} \\
\caption{Performance of region descriptors with regions detected using the  edge weight $1/c(v_1,v_2)$.
Shown are the HKS descriptor (first row) and SI-HKS descriptor (second row).\label{fig_desc_ew3}
}
    \end{center}
\end{small}
\end{figure*}
\fi

%In order to test how discriminative region descriptors are we need to check how much small distance (in the descriptor space) indicates big overlap between regions.
%In order to measure this, for every region in a transformed shape we performed a nearest neighbor search among all null's regions.
In the second experiment, the discriminativity of region descriptors was evaluated by measuring the relation between distance in the descriptor space and
the overlap between the corresponding regions.

Using the notation from the previous section, let $Y_i$ be one of the $n$ maximally stable components detected on a transformed shape $Y$,
$X'_i$ its image on the null shape $X$ under the ground truth correspondence, and let $X_j$ denote one of the $m$ maximally stable components detected on the null shape.
A groundtruth relation between the regions is established by fixing a minimum overlap $\rho=0.75$ and deeming $Y_i$ and $X_j$ matching if $o_{ij} = O(X'_i,X_j) \ge \rho$.
Let us now be given a region descriptor $\beta$; for simplicity we assume the distance between the descriptors to be the standard Euclidean distance.
By setting a threshold $\tau$ on this distance, $Y_i$ and $X_j$ will be classified as positives if $d_{ij} = \| \beta(Y_i) - \beta(X_j) \| \le \tau$.
We define the \emph{true positive rate} as the ratio
\begin{eqnarray}
\mathrm{TPR} &=& \frac{|\{ d_{ij} \le \tau \} |}{|\{ o_{ij} \ge \rho\} |};
\end{eqnarray}
similarly, the \emph{false positive rate} is defined as
\begin{eqnarray}
\mathrm{FPR} &=& \frac{|\{ d_{ij} > \tau \} |}{|\{ o_{ij} < \rho\} |}.
\end{eqnarray}
A related quantity is the \emph{false negative rate} defined as $\mathrm{FNR} = 1-\mathrm{TPR}$. By varying the threshold $\tau$, a set of pairs $(\mathrm{FPR},\mathrm{TPR})$ referred to as the \emph{receiver operator characteristic} (ROC) curve is obtained.
The particular point on the ROC curve for which the false positive and false negative rates coincide is called \emph{equal error rate} (EER).
We use EER as a scalar measure for the descriptor discriminativity. Ideal descriptors have $\mathrm{EER}=0$.

Another descriptor performance criterion used here considers the first matches produced by the descriptor distance.
For that purpose, for each $X_i$ we define its first match as the $Y_{j^*(i)}$ with $j^*(i) = \mathrm{arg}\min_j d_{ij}$ (nearest neighbor of $\beta(X_i)$ in the descriptor space).
The \emph{matching score} is defined as the ratio of correct first matches for a given overlap $\rho$,
\begin{eqnarray}
\mathrm{score}(\rho) &=& \frac{|\{ o_{ij^*(i)} \ge \rho \} |}{m}.
\end{eqnarray}

The following four weighting functions exhibiting best repeatability scores in the previous experiment were used to define region detectors:
the edge weight $1/h_t(v_1,v_2)$ with $t=1024$ (absolute winner in terms of repeatability), the vertex weight $h_t(v,v)$ (second-best repeatability), its edge-weight counterpart $| h_t(v_1,v_1) - h_t(v_2,v_2)|$ (gives lower repeatability scores but supplies almost twice correspondences), and the edge weight $1/c(v_1,v_2)$ (best scale invariant detector).
Given the maximally stable components detected by a selected detector, region descriptors were calculated. We used two types of point descriptors: the heat kernel signature $h_t(v,v)$ sampled at six time values $t=16,22.6,32,45.2,64,90.5,128$, and its scale invariant
version $\hat{h}_\omega(v,v)$, for which we have taken the first six discrete frequencies of the Fourier transform (these are settings identical to \cite{Iasonas}). These point descriptors were used to create region descriptors
using averaging (\label{eq:region-desc-avg}) and local bags of features (\label{eq:region-desc-bof}). Bags of features were tested with vocabulary sizes $p=10$ and $12$.
Table~\ref{table_desc_performance} summarizes the performance in terms of EER of different combinations of weighting functions and region descriptors.
Figure~\ref{fig_ROCs} depicts the ROC curves of different descriptors of vertex- and edge-weighted maximally stable component detectors.

\ifx\isfulltr\undefined
Figures~\ref{fig_desc_vw}--\ref{fig_desc_ew1}
\else
Figures~\ref{fig_desc_vw}--\ref{fig_desc_ew3}
\fi
show the number of correct first matches and the matching score as a function of the overlap for different choices of weighting functions and descriptors.
\ifx\isfulltr\undefined
\else
Examples of matching regions are depicted in Figure~\ref{fig_retrieval}.
\fi

We conclude that the scale invariant HKS descriptor consistently exhibits higher performance in both the average and bag of features flavors. The latter flavors
perform approximately the same. The HKS descriptor, on the other hand, performs better in the bag of feature setting, though never reaching the scores of SIHKS.
Surprisingly, as can be seen from
\ifx\isfulltr\undefined
Figures~\ref{fig_desc_vw}--\ref{fig_desc_ew1},
\else
Figures~\ref{fig_desc_vw}--\ref{fig_desc_ew3},
\fi
the SIHKS descriptor is consistently more discriminative even in transformations not including scaling.

\begin{table*}[tb]
    \begin{small}
        \begin{center}
            \begin{tabular}[width=\columnwidth]{lcccccc}
                \hline
                \hline
                 Weighting & HKS & HKS & HKS & SI-HKS & SI-HKS & SI-HKS   \\
                 function & Avgerage & BoF($p=10$) & BoF($p=12$) & Avgerage & BoF($p=10$) & BoF($p=12$) \\
                \hline
                 $h_t(v,v)$        & 0.311 &  0.273 & 0.278 & 0.093 & 0.091 & 0.086 \\
                 $1/h_t(v_1,v_2)$  & 0.304 &  0.275 & 0.281 & 0.104 & 0.093 & 0.090 \\
                 $| h_t(v_1,v_1) - h_t(v_2,v_2)|$  & 0.213 &  0.212 & 0.222 & 0.085 & 0.091 & 0.094 \\
                 $1/c(v_1,v_2)$  & 0.260 &  0.284 & 0.294 & 0.147 & 0.157 & 0.148 \\
                \hline
                \hline
            \end{tabular}
        \end{center}
    \end{small}
    \caption{Equal error rate (EER) performance of different maximally stable component detectors and descriptors ($t=2048$ was used in all cases). $p$ denotes the vocabulary size in the bag of features region descriptors. }
    \label{table_desc_performance}
\end{table*}

%This is explained by the fact that the SHREC'10 benchmark includes several global and local scale transformations which appear not to influence significantly the repeatability of the detector (even of the scale-dependent ones), yet degrade the discriminativity of the scale-dependent descriptors.

\ifx\isfulltr\undefined
\else
\begin{figure*}[t]
\begin{small}
    \begin{center}
    \includegraphics[width=.9\linewidth,bb=0 0 2427 1344]{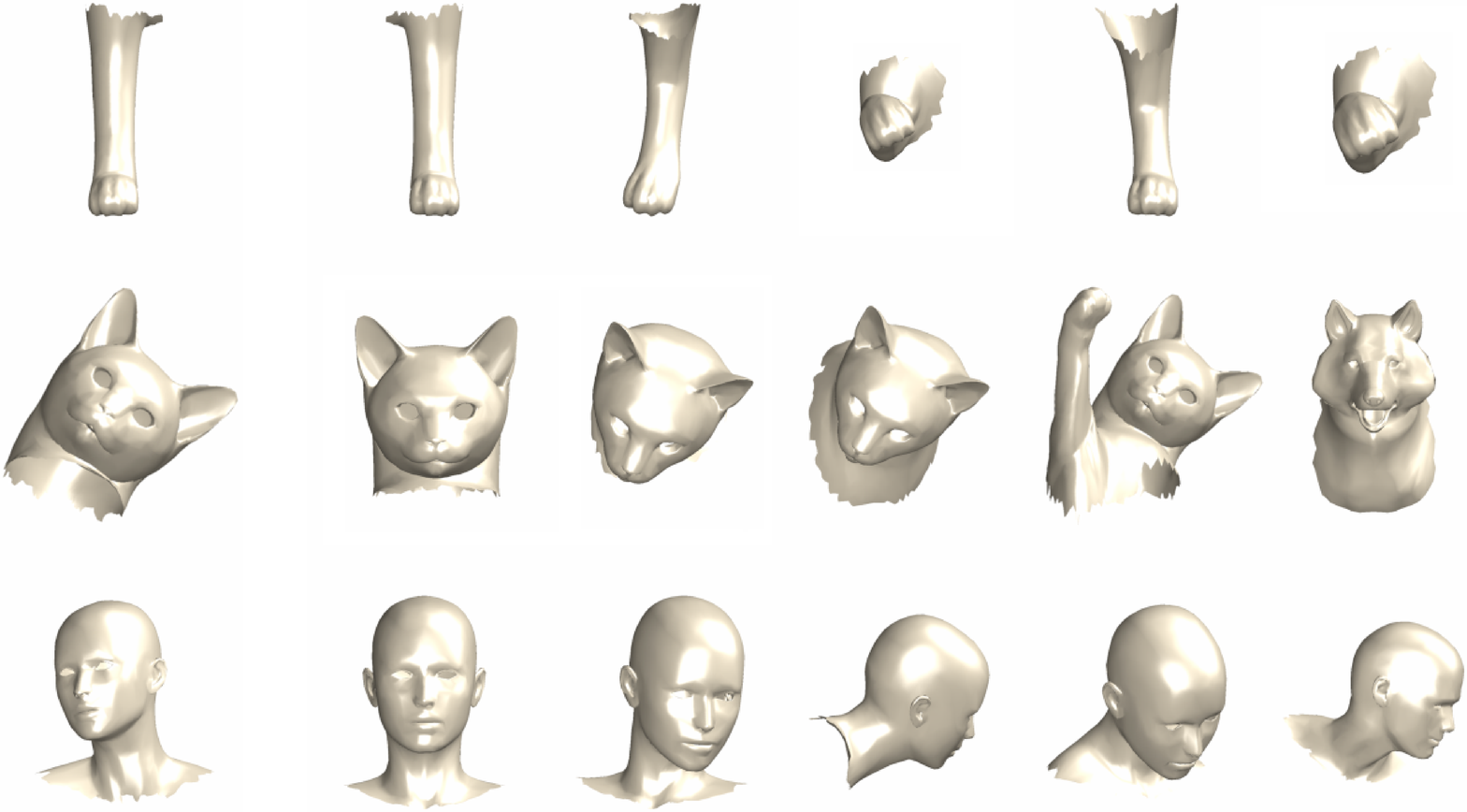} \\
\caption{ Examples of closest matches found for different query regions on the TOSCA dataset.
Shown from left to right are: query, 1st, 2nd, 4th, 10th, and 14th matches.
 Vertex weight $h_t(v,v)$ with $t=2048$ was used as the detector; average SIHKS was
used as the descriptor.
\label{fig_retrieval} }
    \end{center}
\end{small}
\end{figure*}
\fi

\section{Conclusions}

We presented a generic framework for the detection of stable regions in non-rigid shapes. Our approach is based on the maximization of a stability criterion in a component tree representation of the shape with vertex or edge weights. Using diffusion geometric weighting functions allows obtaining a feature detection algorithm that is invariant to a wide class of shape transformations, in particular, non-rigid bending and global scaling, which makes our approach applicable in the challenging setting of deformable shape analysis.
In followup studies, we are going to explore the uses of the proposed feature detectors and descriptors in shape matching and retrieval problems.

%-------------------------------------------------------------------------
%\section{References}
\bibliographystyle{elsarticle-harv}
\bibliography{smi11}

\end{document}